\documentclass[]{bytedance_seed}

\usepackage[toc,page,header]{appendix}

\usepackage{minitoc}
\usepackage{hyperref}
\usepackage{url}
\usepackage{graphicx} 
\usepackage{multicol}
\usepackage{multirow}
\usepackage{booktabs}
\usepackage{listings}
\usepackage{algorithm}
\usepackage{arydshln}
\usepackage[noend]{algpseudocode}
\usepackage{subcaption}
\usepackage{wrapfig}
\usepackage{enumitem}
\usepackage{amsfonts}

\title{INT v.s. FP: A Comprehensive Study of Fine-Grained Low-bit Quantization Formats
}

\author[1,2]{Mengzhao Chen}
\author[3]{Meng Wu}
\author[2]{Hui Jin}
\author[2]{Zhihang Yuan}
\author[2]{Jing Liu}
\author[2]{\\Chaoyi Zhang}
\author[2]{Yunshui Li}
\author[2]{Jie Huang}
\author[2]{Jin Ma}
\author[1]{\\Zeyue Xue}
\author[1]{Zhiheng Liu}
\author[2, \dagger]{Xingyan Bin}
\author[1, \dagger]{Ping Luo}

\affiliation[1]{The University of Hong Kong}
\affiliation[2]{ByteDance Seed}
\affiliation[3]{PicoHeart}

\contribution[\dagger]{Corresponding authors}

\abstract{
Modern AI hardware, such as Nvidia's Blackwell architecture, is increasingly embracing low-precision floating-point (FP) formats to handle the pervasive activation outliers in Large Language Models (LLMs). Despite this industry trend, a unified comparison of FP and integer (INT) quantization across varying granularities has been missing, leaving algorithm and hardware co-design without clear guidance. This paper fills that gap by systematically investigating the trade-offs between FP and INT formats. We reveal a critical performance crossover: while FP excels in coarse-grained quantization, the comparison at fine-grained (block-wise) levels is more nuanced. Our comprehensive comparison demonstrates that for popular 8-bit fine-grained formats (e.g., MX with block size 32), MXINT8 is superior to its FP counterpart in both algorithmic accuracy and hardware efficiency. However, for 4-bit formats, FP (e.g., MXFP4, NVFP4) often holds an accuracy advantage , though we show that NVINT4 can surpass NVFP4 when outlier-mitigation techniques like Hadamard rotation are applied. We also introduce a symmetric clipping method that resolves gradient bias in fine-grained low-bit INT training, enabling nearly lossless performance for MXINT8 training. These findings challenge the current hardware trajectory, demonstrating that a one-size-fits-all FP approach is suboptimal and advocating that fine-grained INT formats, particularly MXINT8, offer a better balance of accuracy, power, and efficiency for future AI accelerators.
}

\date{\today}
\correspondence{ \email{binxingyan@bytedance.com}, \email{pluo@cs.hku.hk}}

\checkdata[Code]{\url{https://github.com/ChenMnZ/INT_vs_FP}}

\begin{document}
\maketitle

%不需要目录就注释掉 注意目录不要和第一页放在一块 要有\newpage
%\newpage
%\tableofcontents
%\newpage

% \input{sections/Abstract}
\section{Introduction}
The proliferation of Large Language Models (LLMs) has been accompanied by a surge in their computational and memory demands~\citep{unveiled}, making quantization an indispensable technique for efficient deployment. A central challenge in quantizing LLMs, particularly those based on the Transformer architecture, is the presence of significant outliers~\citep{massive,llmint8} in activation distributions. These outliers, characterized by their large magnitude but infrequent occurrence, pose a considerable problem for low-precision representations. To accommodate this wide dynamic range, the AI hardware industry~\citep{NVIDIA_Blackwell} is increasingly pivoting towards low-precision floating-point (FP) formats, such as FP8 and FP4. Prominent examples like NVIDIA's Blackwell architecture~\citep{NVIDIA_Blackwell} underscore this trend, favoring the superior dynamic range of FP to handle outliers more gracefully than traditional integer (INT) formats.

% Despite this industry-wide momentum, the comparative advantages of FP and INT formats have not been fully explored under a unified framework. Most existing studies~\citep{smoothquant,prefixquant,spinquant} either focus on a single format in isolation or conduct comparisons at a fixed, coarse granularity (e.g., per-channel quantization). This leaves a critical question unanswered: how does the performance trade-off between INT and FP evolve as the quantization granularity becomes finer? As block-wise quantization—which partitions tensors into smaller blocks and quantizes each independently—becomes a standard technique for mitigating the impact of outliers, understanding its interplay with the underlying number format is paramount for both algorithmic and hardware co-design.

%
However, this industry-wide momentum towards FP formats is based on an incomplete picture. The comparative advantages of FP and INT have not been systematically evaluated across different quantization granularities in a unified framework. Most studies~\citep{smoothquant,prefixquant,spinquant} focus on a single format or compare them only at coarse granularities (e.g., per-channel), failing to answer a critical question: how does the performance trade-off between INT and FP evolve as granularity becomes finer? Since fine-grained (block-wise) quantization is now a standard technique~\citep{ocp,nv_quant_doc} for mitigating outliers, understanding its interaction with the underlying number format is essential for effective algorithm-hardware co-design.

In this paper, we conduct a comprehensive, systematic comparison of fine-grained INT and FP quantization. Our investigation reveals a critical "crossover point" in performance. While FP formats hold a distinct advantage in coarse-grained scenarios, we find that INT formats become highly competitive as the block size shrinks, though the benefit depends heavily on the bit width. As granularity becomes finer, the local dynamic range within each block is reduced, allowing the uniform precision of INT formats to become more effective. This trend is analyzed across modern block-wise formats, such as the 32-element blocks in Microscaling (MX) formats or the 16-element blocks in NVIDIA's (NV) formats. To enable a direct comparison, we introduce and evaluate integer variants (e.g., MXINT8, MXINT6, MXINT4, NVINT4) alongside their standard FP counterparts (e.g., MXFP8, MXFP6, MXFP4, NVFP4).

Our key contributions are as follows:

\begin{itemize}
    \item We develop a theoretical and statistical framework that models the quantization signal-to-noise ratio (QSNR) for both INT and FP formats. This framework enables a direct theoretical comparison of their performance trade-offs and clarifies the crossover points and .
    
    \item We demonstrate that MXINT8 consistently outperforms MXFP8 in both direct-cast inference and low-bit training. We also show that NVINT4 can surpass NVFP4 when combined with Hadamard rotation. Critically, we introduce a symmetric clipping method that resolves a gradient bias, enabling nearly lossless MXINT8 low-bit training.
    
    \item We present a comparative hardware cost analysis, demonstrating that fine-grained INT formats are significantly more area and energy-efficient than their floating-point counterparts at matched throughput.
    
    \item Collectively, our findings challenge the prevailing FP-centric trajectory in AI hardware design and advocate for prioritizing fine-grained INT formats to achieve a more optimal balance of accuracy and efficiency in future AI accelerators.
\end{itemize}

\section{Preliminaries}\label{sec:preliminaries}

Quantization maps a high-precision tensor $\mathbf{X}$ to a lower bit-width. In this section, we present low-bit integer (INT) quantization, floating-point (FP) quantization, quantization granularity with a focus on fine-grained block-wise schemes, and an overview of existing low-bit block formats.

\subsection{Low-Precision Integer Formats}
For $b$-bit integer quantization, we define:
\begin{equation}\label{eq:int_quant}
\mathbf{X_q} = \text{clip}\left(\left\lfloor \frac{\mathbf{X}}{s} \right\rceil, Q_{\min}, Q_{\max} \right) \cdot s,
\end{equation}
where $s$ is the scale factor that normalizes $\mathbf{X}$ to the target integer range, $\lfloor \cdot \rceil$ is round-to-nearest, and $\mathbf{X_q}$ is the dequantized tensor. The clipping ensures that the integer values lie in $[Q_{\min}, Q_{\max}]$ (e.g., for signed $b$-bit integers, $Q_{\min}=-2^{b-1}$ and $Q_{\max}=2^{b-1}-1$).

\subsection{Low-Precision Floating-Point Formats}
Floating-point representation~\citep{ieee_float} uses three fields: the sign bit ($S$), the exponent ($E$), and the mantissa ($M$). We denote a format as E$x$M$y$, where $x$ and $y$ are the numbers of exponent and mantissa bits. The sign determines the polarity, the exponent sets the dynamic range, and the mantissa sets the precision. A floating-point number decodes as:
\begin{equation}
\mathbb{C}_\text{FP} = \begin{cases}
(-1)^{s} \times (1.m)_2 \times 2^{e-\text{bias}} & \text{if } e \neq 0 \text{ (Normal)}, \\
(-1)^{s} \times (0.m)_2 \times 2^{1-\text{bias}} & \text{if } e = 0,\,m \neq 0 \text{ (Subnormal)},
\end{cases}
\end{equation}
where $s$, $e$, and $m$ are the sign, exponent and mantissa values of a float-point number. Hence, $\mathbb{C}_\text{FP}$ denotes the set of representable low-bit floating-point values. Floating-point quantization is:
\begin{equation}\label{eq:fp_quant}
\mathbf{X_q} = \text{Nearest}\!\left(\frac{\mathbf{X}}{s}, \mathbb{C}_\text{FP}\right) \cdot s,
\end{equation}
where $\text{Nearest}(\cdot, \mathbb{C}_\text{FP})$ maps normalized values to the nearest element of $\mathbb{C}_\text{FP}$. Eq.~(\ref{eq:fp_quant}) is a general quantization form that also recovers integer quantization by replacing $\mathbb{C}_\text{FP}$ with $\mathbb{C}_\text{INT}$.

\subsection{Quantization Granularity}
Quantization granularity specifies how scale factors apply across a tensor. Finer granularity usually improves accuracy but increases compute and memory overhead due to more scale factors. Common choices are:
(i) Per-tensor: a single scale for the entire tensor.
(ii) Per-channel: a scale per channel, broadcast along a chosen axis.
(iii) Block-$k$: the tensor is partitioned into $1 \times k$ blocks along one dimension, and each block has its own scale.
Block quantization is a key technique for improving accuracy at low precision. In this paper, we mainly focus on block quantization.

\begin{table}[]
    \centering
    \caption{Low-bit formats name and their correspond represented range and scale factors.}
    \label{tab:low_bit_formats}
    \begin{tabular}{lcccccc}
    \hline
    Format & Block Size & Max Value & Min Value & Dynamic Range & Scale-1 & Scale-2  \\
    \hline
    MXFP8 (E4M3) & 32 & $\pm 448 $ & $\pm2^{-9}$ & $1.75 \times 2^{17}$ & UE8M0 & -  \\
    \hline
    MXINT8 & 32 & $127$ & $1$ & $127$ & UE8M0 & - \\
    \hline
    MXFP6 (E2M3) & 32 & $\pm 7.5 $ & $\pm 0.125$ & $60$ & UE8M0 & -  \\
    \hline
    MXINT6 & 32 & $\pm 31 $ & $\pm 1$ & $31$ & UE8M0 & -  \\
    \hline
    MXFP4 (E2M1) & 32 & $\pm 6 $ & $\pm 0.5$ & $12$ & UE8M0 & -  \\
    \hline
    MXINT4 & 32 & $\pm 7 $ & $\pm 1$ & $7$ & UE8M0 & -  \\
    \hline
    NVFP4 & 16  & $\pm 6 $ & $\pm 0.5$ & $12$ & E4M3 & FP32  \\
    \hline
    NVINT4 & 16 & $\pm 7 $ & $\pm 1$ & $7$ & E4M3 & FP32  \\
    \hline
    \end{tabular}
\end{table}

\subsection{Block-Quantization Formats}
To improve low-bit accuracy, OCP~\citep{ocp} proposes the Microscaling (MX) format, which uses a shared UE8M0\footnote{UE8M0 is an 8-bit unsigned floating-point format with eight exponent bits and zero mantissa bits.} scale for each block of 32 elements. This fine-grained scaling reduces quantization error. Recently, NVIDIA Blackwell-series GPUs~\citep{nv_quant_doc} provide native hardware support for MXFP8/MXFP6/MXFP4. Traditionally, FP8 has E4M3 and E5M2 variants, and FP6 has E2M3 and E3M2 variants. We consider E4M3 for MXFP8 and E2M3 for MXFP6 because mantissa bits are more critical to the performance of fine-grained quantization, consistent with prior work~\citep{liu2024deepseek,mxfp8_training,ocp}.
Furthermore, NVIDIA proposes NVFP4, which enhances MXFP4 by reducing the block size from 32 to 16 and replacing the UE8M0 scale with an E4M3 scale. NVFP4 also introduces a second-level per-tensor scale to prevent overflow of the first-level E4M3 scale.
Therefore, current hardware tends to support low-bit fine-grained floating-point formats. To enable fair comparison between low-bit floating-point and integer formats, we also introduce four corresponding integer variants: MXINT8, MXINT6, MXINT4, and NVINT4.
Details of these low-bit formats are listed in Table~\ref{tab:low_bit_formats}.

\begin{figure}[t]
    \centering
    \begin{minipage}{0.45\textwidth}
        \centering
        \includegraphics[width=\linewidth]{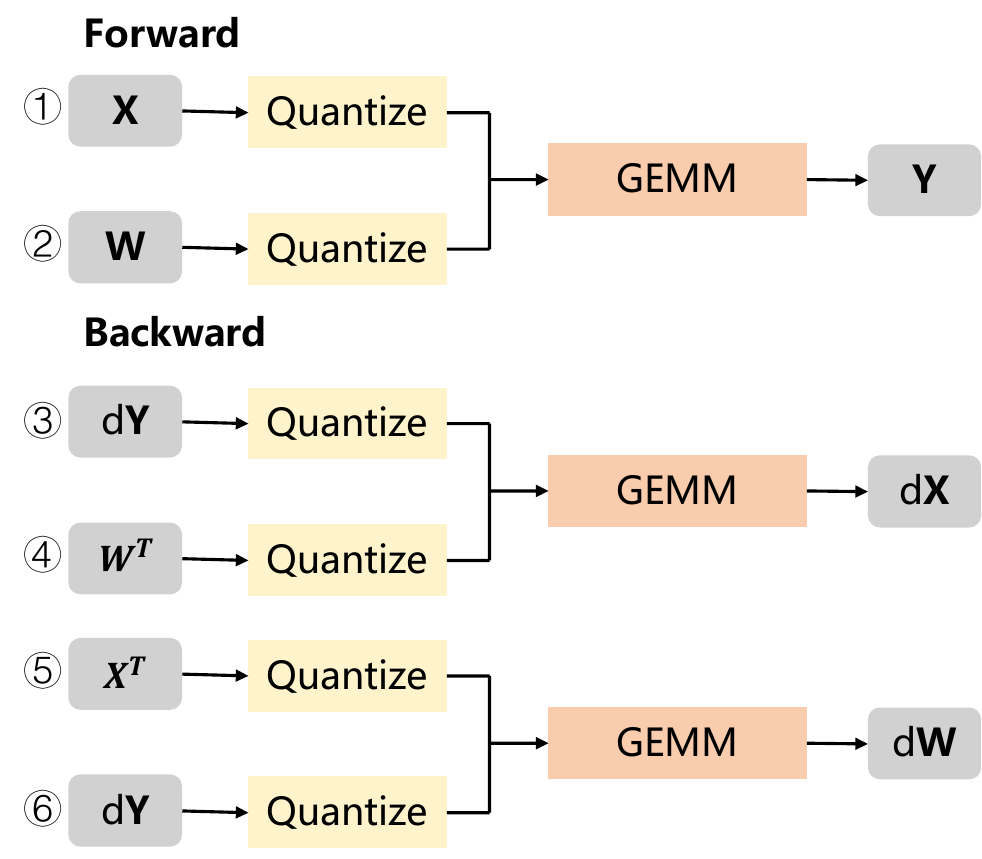}
        \captionof{figure}{Compute flow of low-bit forward and backward propagation of linear layer.}
        \label{fig:quant_flow}
    \end{minipage}
    \hfill
    \begin{minipage}{0.48\textwidth}
        \centering
        \includegraphics[width=\linewidth]{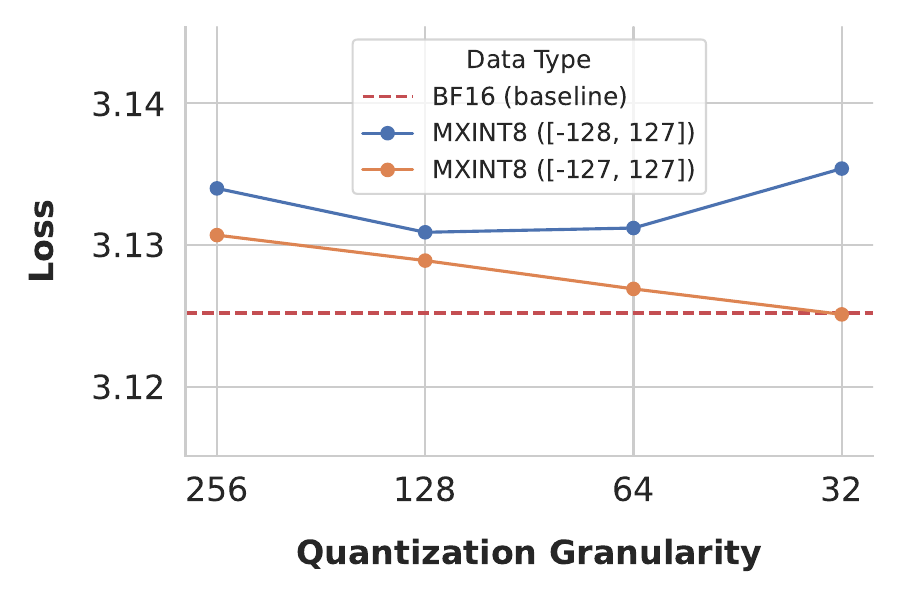}
        \captionof{figure}{Impact of clipping range on INT8 final training loss on 145M model with 20B training tokens. Scale factor is kept on BF16 to emphasize the harm of asymmetric representation space during low-bit training.}
        \label{fig:int_range_ablation}
    \end{minipage}
\end{figure}

\section{Quantization Recipe}~\label{sec:recipe}
This section illustrates the computation flow for low-bit inference and training in Sec.~\ref{sec:compute_flow}, and details the scale-factor computation used in quantization in Sec.~\ref{sec:quantization_operation}.
\subsection{Quantization Compute Flow}\label{sec:compute_flow}
Figure~\ref{fig:quant_flow} shows an example of using low-bit GEMM in a linear layer during forward and backward propagation. Given high-precision (e.g., BFloat16) activations $\mathbf{X}$ and weights $\mathbf{W}$, the forward pass of the quantized linear layer\footnote{We omit the bias term.} is:
\begin{equation}\label{eq:forward_Y}
    \mathbf{Y}
    =
    \underbrace{\text{Quantize}(\mathbf{X})}_{\textcircled{1}}
    \,
    \underbrace{\text{Quantize}(\mathbf{W})}_{\textcircled{2}}.
\end{equation}
The backward pass to compute $d\mathbf{X}$ and $d\mathbf{W}$ is:
\begin{align}
    d\mathbf{X} &=
        \underbrace{\text{Quantize}(\mathbf{dY})}_{\textcircled{3}}
        \,
        \underbrace{\text{Quantize}(\mathbf{W}^{T})}_{\textcircled{4}},
        \label{eq:backward_dx}\\
    d\mathbf{W} &=
        \underbrace{\text{Quantize}(\mathbf{X}^{T})}_{\textcircled{5}}
        \,
        \underbrace{\text{Quantize}(\mathbf{dY}^{T})}_{\textcircled{6}}.
        \label{eq:backward_dw}
\end{align}
$\text{Quantize}(\cdot)$ maps high-precision tensors to low-bit representations. Thus, there are six quantization operations in one linear layer: \textcircled{1} $\mathbf{X}$ and \textcircled{2} $\mathbf{W}$ in Eq.~(\ref{eq:forward_Y}); \textcircled{3} $\mathbf{dY}$ and \textcircled{4} $\mathbf{W}^T$ in Eq.~(\ref{eq:backward_dx}); \textcircled{5} $\mathbf{X}^T$ and \textcircled{6} $\mathbf{dY}^T$ in Eq.~(\ref{eq:backward_dw}). Block-wise quantization requires tensors to be quantized along the GEMM reduction dimension to gain hardware benefits. Therefore, \textcircled{1} and \textcircled{5}, \textcircled{2} and \textcircled{4}, and \textcircled{3} and \textcircled{6} are quantized along different axes~\citep{liu2024deepseek,qsnr}. We separately analyze the quantization error of these six operations in Sec.~\ref{sec:tensor}.

\subsection{Quantization Operation}\label{sec:quantization_operation}
\textbf{UE8M0 scale factor.} The scale factor $s$ in Eq.~(\ref{eq:int_quant}) and Eq.~(\ref{eq:fp_quant}) is computed with the AbsMax quantizer:
\begin{equation}\label{eq:bf16_scales}
    s = \frac{\text{AbsMax}(\mathbf{X})}{Q_{max}},
\end{equation}
where $\text{AbsMax}(\mathbf{X})$ is the maximum absolute value within the group of values that share a single scale factor, and $Q_{max}$ is the maximum value of the quantized type (see Table~\ref{tab:low_bit_formats}). Eq.~(\ref{eq:bf16_scales}) maps the largest magnitude in high precision to the maximum representable low-precision value without clipping. OCP~\citep{ocp} further converts the high-precision scale factor to the UE8M0 format for MX formats:
\begin{equation}\label{eq:ocp_e8m0}
    s' = 2^{\text{clip}\!\left(\left\lfloor\log_2(\text{AbsMax}(\mathbf{X}))\right\rfloor - \left\lfloor\log_2(Q_{max})\right\rfloor, -127, 127\right)},
\end{equation}
where $\lfloor \cdot \rfloor$ denotes rounding down. Eq.~(\ref{eq:ocp_e8m0}) rounds the high-precision scale down to the nearest UE8M0 value, which introduces extra clipping error. Following existing works~\citep{mxfp4_training,mxfp4_vit,mxfp8_training}, we round up the UE8M0 scale based on Eq.~(\ref{eq:bf16_scales}) to avoid this error:
\begin{equation}\label{eq:our_e8m0}
    s' = 2^{\text{clip}\!\left(\left\lceil\log_2(s)\right\rceil, -127, 127\right)},
\end{equation}
where $\lceil\cdot\rceil$ denotes rounding up.

\textbf{Symmetric Clipping.} Floating-point formats are naturally symmetric around zero. In contrast, signed integers in two's complement have one extra negative value: for a $b$-bit integer, $Q_{min} = -2^{b-1}$ and $Q_{max} = 2^{b-1} - 1$~\citep{nv_quant_doc}. We find that this asymmetric range usually does not affect inference. However, as shown in Figure~\ref{fig:int_range_ablation}, it degrades INT8 training due to a persistent negative bias in gradients. Finer-grained quantization suffers more because more values fall into the unique negative endpoint $Q_{min}$. For INT8, the minimum value in a group can still map to $-128$ even when the scale is set to $\text{AbsMax}(\mathbf{X})/127$ due to BFloat16 arithmetic precision (see Sec.~\ref{sec:appendix_sym_clip} for details).
Therefore, we use a symmetric integer range for all INT quantizers as shown in Table~\ref{tab:low_bit_formats}:
\[
Q_{min} = -(2^{b-1}-1), \quad Q_{max} = 2^{b-1}-1,
\]

In this section, we analyze low-bit integer and floating-point formats and build a theoretical framework for comparing them. Section~\ref{sec:theoretical_qsnr} derives theorems for the quantization signal-to-noise ratio (QSNR), and Section~\ref{sec:theoretical_comparison} compares low-bit formats based on the theoretical QSNR.

\section{Theoretical Framework}\label{sec:theoretical_analysis}

\subsection{Theoretical QSNR}~\label{sec:theoretical_qsnr}
\textbf{QSNR Metric.} We use the Quantization Signal-to-Noise Ratio (QSNR, dB)~\citep{qsnr} to measure numerical fidelity under different quantization schemes. QSNR is the ratio of the power of the original signal $\mathbf{X}$ to the power of the quantization noise $\mathbf{X}-\mathbf{X}_q$, expressed in decibels:
\begin{equation}
\mathrm{QSNR} = -10 \log_{10}\!\left(\frac{\lVert \mathbf{X}-\mathbf{X}_q \rVert^{2}}{\lVert \mathbf{X} \rVert^{2}}\right).
\end{equation}
A higher QSNR means the quantized vector better preserves the magnitude and direction of the original vector.

\begin{figure}
    \centering
    \includegraphics[width=0.7\linewidth]{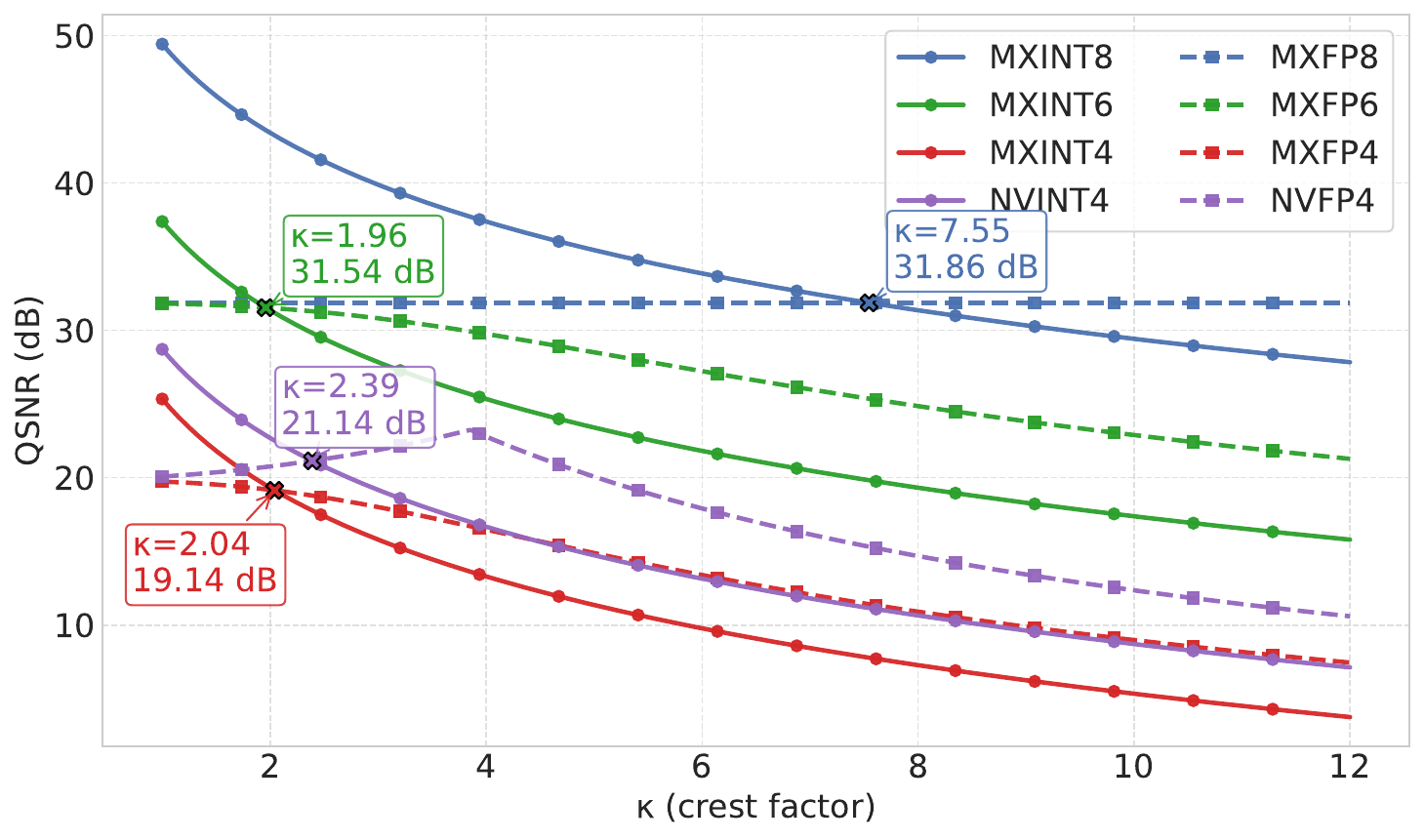}
    \caption{Theoretical QSNR comparison between various integer (INT) and floating-point (FP) formats across a range of crest factors ($\kappa$), derived from Eq.~(\ref{eq:int_qsnr}) and Eq.~(\ref{eq:fp_qsnr}). The boxes represent the crest factor and QSNR of the crossover point of the INT and FP curves.}
    \label{fig:qsnr_vs_kappa}
\end{figure}

% \subsection{Theoretical Analysis}

\textbf{Common assumptions.}
We consider block vectors $\mathbf{X}\in\mathbb{R}^k$ with i.i.d.\ entries $X_i\sim\mathcal{N}(0,\sigma^2)$. The block root-mean-square (RMS) equals $\sigma$, and the \textbf{crest factor} is
\begin{equation}\label{eq:crest_factor}
\kappa := \frac{\max(|\mathbf{X}|)}{\sigma}.
\end{equation}
We use blockwise absolute-maximum (AbsMax) scaling:
\begin{equation}
    s'=\rho\,s,
\end{equation}
where $s$ is the high-precision scale from Eq.~(\ref{eq:bf16_scales}), and $\rho$ models the overhead of the low-precision scale. For example, the UE8M0 scale in Eq.~(\ref{eq:our_e8m0}) has $\rho\in[1,2)$, while for the E4M3 scale in NV-format we set $\rho=1$ since it is close to BFloat16 scales.

\textbf{Theorem 1 (INT QSNR).}
Under $b$-bit INT quantization, the QSNR (in dB) is
\begin{equation}\label{eq:int_qsnr}
\mathrm{QSNR_{INT}} \approx
\begin{cases}
4.78 \;+\; 6.02\,b \;-\; 20\log_{10}(\rho) \;-\; 20\log_{10}(\kappa), & \text{UE8M0 scale} \\[6pt]
4.78 \;+\; 6.02\,b \;-\; 20\log_{10}(\kappa) \;+\; 10\log_{10}\!\left(\dfrac{g}{g-1}\right), & \text{E4M3 scale}
\end{cases}
\end{equation}
A detailed proof of Theorem 1 appears in Sec.~\ref{sec:theorem1_proof}, where $b$ is the bit width, $\rho$ is the scale overhead, $\kappa$ is the crest factor in Eq.~(\ref{eq:crest_factor}), and $g$ is the block size.

\textbf{Interpretation of Theorem 1.}
(i) Each extra bit gives $\approx 6.02$ dB. 
(ii) UE8M0 scaling incurs up to $20\log_{10}(\rho)\le 6.02$ dB loss. 
(iii) A larger crest factor $\kappa$ reduces QSNR; smaller blocks usually reduce $\kappa$ and improve QSNR.
(iv) E4M3 scaling has no $\rho$ overhead and avoids the per-block maximum error, giving a $10\log_{10}\!\left(\dfrac{g}{g-1}\right)$ QSNR gain.

\textbf{Theorem 2 (FP QSNR).}
Under FP quantization, the QSNR (in dB) is
\begin{equation}\label{eq:fp_qsnr}
\mathrm{QSNR_{FP}} \approx
\begin{cases}
-10\log_{10}\!\left(\alpha_M \, w_{\mathrm{norm}} \;+\; \beta \,(\rho\,\kappa)^2 \, p_{\mathrm{sub}}\right), & \text{UE8M0 scale} \\[6pt]
-10\log_{10}\!\left(\alpha_M \, \big(w_{\mathrm{norm}} - \tfrac{\kappa^2}{g}\big) \;+\; \beta \,\kappa^2 \, p_{\mathrm{sub}}\right), & \text{E4M3 scale}
\end{cases}
\end{equation}

A detailed proof of Theorem 2 appears in Sec.~\ref{sec:theorem2_proof}, with $\alpha_M=\frac{1}{24\cdot 2^{2M}}$ (mantissa resolution term) and $\beta=\frac{2^{2(1-B-M)}}{12\,Q_{\max}^2}$. Here $M$ is the mantissa bit width, $B$ is the exponent bias, and $Q_{\max}$ is the largest finite normal magnitude of the target FP format (e.g., $Q_{\max}=448$ for E4M3). The terms $w_{\mathrm{norm}}$ and $p_{\mathrm{sub}}$ measure how much of the distribution falls into the normal and subnormal regions (after scaling): $w_{\mathrm{norm}}$ is the fraction of signal energy carried by normal FP numbers and incurs mantissa quantization error $\alpha_M$; $p_{\mathrm{sub}}$ is the probability that a value encodes as subnormal and incurs a fixed absolute step error.

\textbf{Interpretation of Theorem 2.}
(i) The mantissa bit width sets the upper bound on FP QSNR. With ample dynamic range ($w_{\mathrm{norm}}\approx 1$ and $p_{\mathrm{sub}}\approx 0$),
$\mathrm{QSNR}\approx 13.80 + 6.02\,M$ dB, independent of block granularity and the distribution of $\mathbf{X}$.
(ii) A larger crest factor $\kappa$ increases the share of subnormals and reduces QSNR. Finer-grained blocks reduce $\kappa$, lower $p_{\mathrm{sub}}$, and improve QSNR.
(iii) E4M3 scaling has no $\rho$ overhead and avoids the per-block maximum error, reducing $\frac{\kappa^2}{g}$ error energy in the normal region.

\subsection{Theoretical Comparisons}~\label{sec:theoretical_comparison}
With Eq.~(\ref{eq:int_qsnr}) in Theorem 1 and Eq.~(\ref{eq:fp_qsnr}) in Theorem 2, we estimate the QSNR of low-bit integer and floating-point formats for a given bit width and target distribution (via $\kappa$). Specifically, we set $\rho=1.5$ to imitate UE8M0 scale. As shown in Figure~\ref{fig:qsnr_vs_kappa}, we observe:
\begin{itemize}
    \item \textbf{MXINT8 \emph{vs.} MXFP8}: MXFP8 QSNR varies smoothly due to its ample dynamic range. MXINT8 outperforms FP8 when $\kappa < 7.55$.
    \item \textbf{MXINT6 \emph{vs.} MXFP6}: MXFP6 has the same QSNR as MXFP8 at small $\kappa$, because both MXFP6 and MXFP8 have three mantissa bits. However, FP6 QSNR decreases rapidly as $\kappa$ increases due to limited dynamic range. MXINT6 outperforms MXFP6 only when $\kappa < 1.96$.
    \item \textbf{MXINT4 \emph{vs.} MXFP4}: MXINT4 outperforms MXFP4 when $\kappa < 2.04$.
    \item \textbf{NVINT4 \emph{vs.} NVFP4}: NVINT4 outperforms NVFP4 when $\kappa < 2.39$. One interesting phenomenon is that NVFP4's QSNR even increase when $\kappa < 4$, this can be explained by Eq~(\ref{eq:fp_qsnr}) that larger $\kappa$ can decrease the error of normal domain but increase the error of subnormal domain. In the relatively small $\kappa$ ($\kappa < 4$), normal domain dominate the error so that NVFP4' QSNR can increase when $\kappa < 4$.
\end{itemize}

Therefore, the key factor when comparing FP and INT formats is the data’s crest factor $\kappa$.

% Furthermore, Figure~\ref{fig:crest_factor} shows that the crest factor $\kappa$ decreases as block size decreases. For a Gaussian distribution, $\kappa$ decreases from $3.449$ at block size $1024$ to $2.365$ at block size $32$. For an outlier-heavy distribution, $\kappa$ decreases from $8.054$ at block size $1024$ to $3.048$ at block size $32$. Since $3.048$ lies below the $8$-bit/$4$-bit intersection in Figure~\ref{fig:qsnr_vs_kappa}, this explains why MXINT8, MXINT4, and NVINT4 outperform their floating-point counterparts in most cases, as shown in Figure~\ref{fig:inference_kl_a}. However, MXINT6 lags behind MXFP6 because it only outperforms when $\kappa < 2.04$, whereas even a Gaussian distribution has $\kappa = 2.365 > 2.04$ at block size $32$. In addition, the benefit of integer quantization increases as $\kappa$ decreases, so outlier-alleviation techniques~\citep{quarot,spinquant,omniquant,prefixquant} can further improve integer performance relative to floating-point quantization as demonstrated in Figure~\ref{fig:inference_kl_b}.

% \input{figure_tex/qsnr}

%This explains the strong performance of blockwise INT8, the robustness of FP6 over INT6 except at very fine granularity, and the competitiveness of INT4 relative to FP4 for most tensors at moderate-to-small block sizes.

\section{FP \emph{v.s.} INT}\label{sec:fp_vs_int}
We compare low-bit integer and floating-point formats at three levels. Section~\ref{sec:tensor} analyzes the crest factor and QSNR for six types of intermediate tensors in Figure~\ref{fig:quant_flow}, offering a tensor-level perspective. Section~\ref{sec:inference} evaluates direct-cast inference, quantizing only the forward process. Section~\ref{sec:training} presents results for low-bit training, quantizing both forward and backward processes.

\subsection{Tensor-wise Analysis}\label{sec:tensor}

\textbf{Setup.} To measure the QSNR in real data, we feed 8 WikiText2~\citep{wikitext2} sequences of length 4096 into Llama-3.1-8B, run both forward and backward propagation in BFloat16 precision, and capture the six intermediate tensors (weights, activations, and gradients) indicated by \textcircled{1}--\textcircled{6} in Figure~\ref{fig:quant_flow}. Llama-3.1-8B contains 224 linear layers across all transformer blocks. We collect these tensors for all 224 linear layers, leads totally $224\times6=10752$ tensors, and use them to compute the crest factors under different block size and QSNR under different low-bits formats. Specifically, QSNR is directly calculated tensor-wise, and crest factor is calculated block-wise and than average across the tensor. Additonally, we also apply random hadamard rotation~\cite{quarot} with dimension as $32\times32$ to measure the effectiveness of such outlier surpression technical to crest factor and QSNR.

% Because tensors of the same type have similar distributions, we report the QSNR averaged within each of the six types separately. Specifically, we evaluate INT and FP quantization at 8, 6, and 4 bits. We use E8M0 scales for MX-format at 8 and 6 bits, and E4M3 scales for NV-format at 4 bits, since NV-format significantly outperforms MX-format in the 4-bit setting.

\begin{figure}[!t]
    \centering
    \begin{subfigure}[b]{0.9\linewidth}
        \centering
        \includegraphics[width=\linewidth]{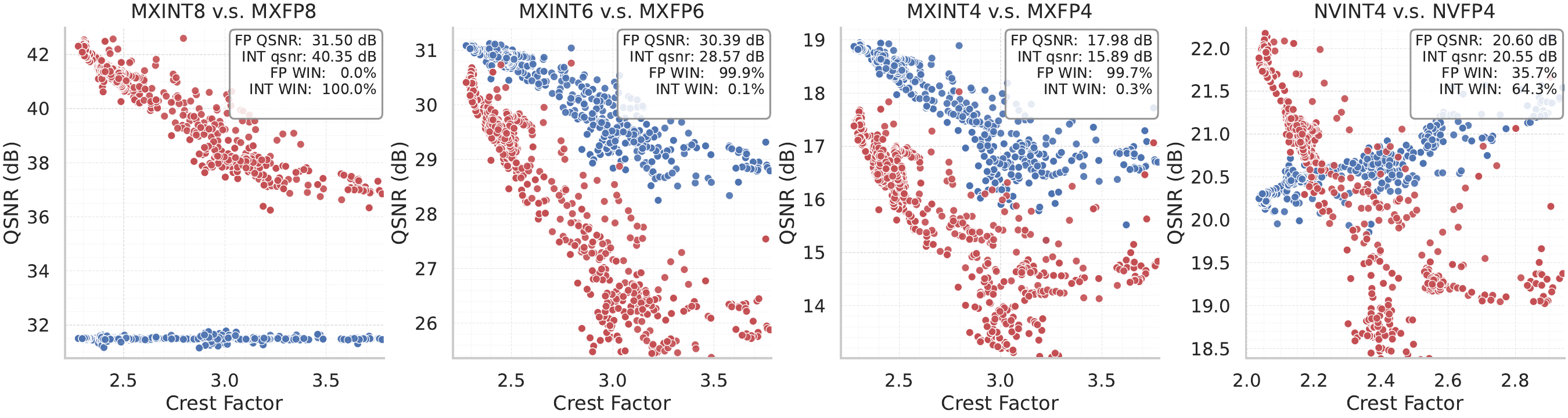}
        \caption{QSNR across crest factor}\label{fig:qsnr}
    \end{subfigure}
    \begin{subfigure}[b]{0.9\linewidth}
        \centering
\includegraphics[width=\linewidth]{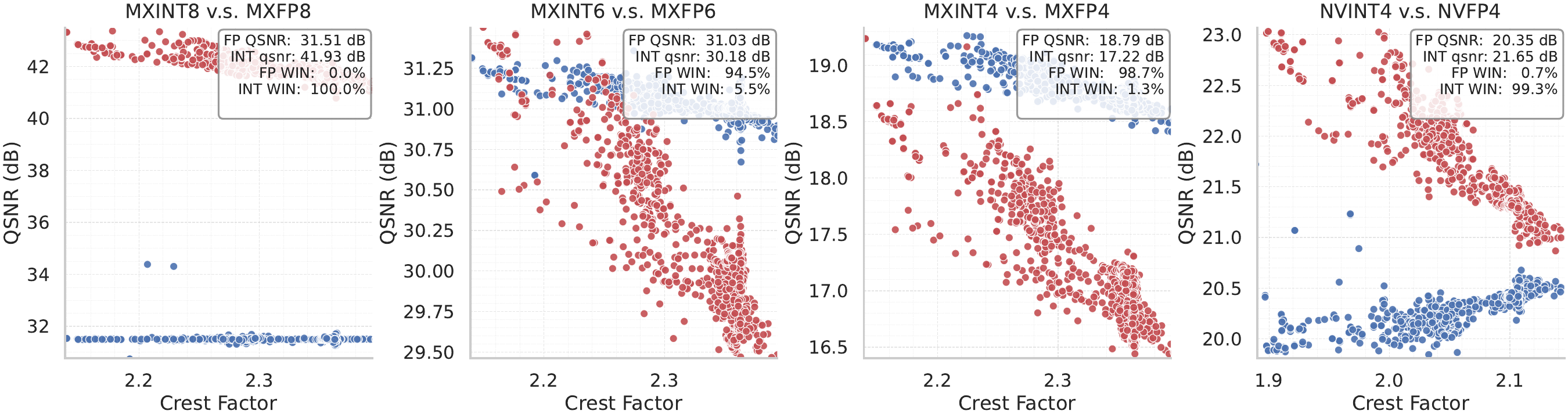}
        \caption{QSNR across crest factor (w/ Hadamard rotation)}\label{fig:qsnr_rotation}
    \end{subfigure}
    \caption{Practical QSNR across crest factors from 10752 tensors source from \textcircled{1} to \textcircled{6} in compute flow in Figure~\ref{fig:quant_flow}. (a) is the results from vanilla tensor and (b) applies random hadamard rotation to the tensor before quantization. The box in top right report the average QSNR of INT and FP quantization, and the win rates of INT and FP quantization.}
    \label{fig:qsnr_all}
\end{figure}
% \usepackage{multirow}
\begin{table}[]
\centering
\caption{Summary statistics of the crest factor by block size in boxplot form. Q1 and Q3 denote the 25\% and 75\% quantiles, respectively.}~\label{tab:crest_factor}
\begin{tabular}{ccccccc}
\hline
Type                                                & Block Size & Min  & Q1   & Median & Q3    & Max   \\
\hline
\multirow{3}{*}{Crest factor}                       & -1         & 3.55 & 4.26 & 6.2    & 11.97 & 60.15 \\
                                                    & 32         & 2.28 & 2.40  & 2.48   & 2.96  & 4.26  \\
                                                    
                                                    & 16         & 2.04 & 2.13 & 2.16   & 2.39  & 3.16  \\
                                                    \hline
\multirow{3}{*}{Crest factor w/ hadamard rotatioin} & -1         & 3.62 & 3.9  & 4.15   & 5.79  & 13.02 \\
                                                    & 32         & 1.91 & 2.29 & 2.35   & 2.36  & 2.57  \\
                                                    & 16         & 1.77 & 2.06 & 2.1    & 2.11  & 2.21  \\
                                                    \hline
\end{tabular}
\end{table}

\textbf{Crest factor results.} Table~\ref{tab:crest_factor} reports crest factor statistics in boxplot form. We focus on the \textbf{75\% quantile (\emph{i.e.}, Q3)}, which reflects typical worst-case behavior across 75\% of cases.
For channel-wise quantization (block size $-1$), Q3 is $11.97$, which is far above the crossover point in Figure~\ref{fig:qsnr_vs_kappa}. This indicates that FP outperforms INT in most cases with coarse granularity.
For the MX-format with block size $32$, Q3 is $2.96$. This value is well below the MXINT8 \emph{v.s.} MXFP8 crossover point ($7.55$), so MXINT8 outperforms MXFP8 in most cases. In contrast, $2.96$ is above the MXINT6 \emph{v.s.} MXFP6 and MXINT4 \emph{v.s.} MXFP4 crossover points ($1.96$ and $2.04$), so MXINT6 and MXINT4 underperform their FP counterparts. After Hadamard rotation, Q3 decreases from $2.96$ to $2.39$, which remains below $7.55$ but above $1.96$ and $2.04$; thus, MXINT8 still wins, while MXINT6 and MXINT4 still lag behind MXFP6 and MXFP4.
For the NV-format with block size $16$, Q3 is $2.39$, which equals the NVINT4 \emph{v.s.} NVFP4 crossover point and then decreases to $2.11$ after Hadamard rotation, favoring NVINT4 over NVFP4 post-rotation.

\textbf{Crest factor v.s. QSNR results.}
Figure~\ref{fig:qsnr_all} reports measured QSNR across crest factors. The empirical trends closely follow the theoretical comparisons in Sec.~\ref{sec:theoretical_analysis} (Theorems~1--2) and the aforementioned crest factor reults:

\begin{itemize}[nosep,leftmargin=*,labelsep=0.5em]
    \item \textbf{MXINT8 \emph{v.s.} MXFP8}: The QSNR of MXFP8 is nearly constant at $31.50$ because of its large dynamic range and mantissa-bit bound. MXINT8 has an average QSNR of $40.35$, and thus significantly outperforms MXFP8.
    \item \textbf{MXINT6 \emph{v.s.} MXFP6 and MXINT4 \emph{v.s.} MXFP4}: MXINT6 and MXINT4 consistently lag behind MXFP6 and MXFP4, with or without random Hadamard rotation. 
    \item \textbf{NVINT4 \emph{v.s.} NVFP4}: Although the win rate of NVINT4 is $64.3\%$, its average QSNR is $20.55$, which is slightly below NVFP4's $20.60$ because NVINT4's QSNR decreases faster than NVFP4's as the crest factor increases. After random Hadamard rotation, NVINT4's average QSNR rises to $21.65$, surpassing NVFP4's $20.35$. Note that NVFP4's QSNR decreases from $20.60$ to $20.35$ after rotation, which is consistent with Figure~\ref{fig:qsnr_vs_kappa}: rotation reduces the crest factor, and when the crest factor is below $4$, NVFP4's QSNR increases with the crest factor, so a reduction in crest factor lowers its QSNR.
\end{itemize}

Overall, real-data measurements corroborate the theory in Sec.~\ref{sec:theoretical_analysis}.

%This explains the strong performance of blockwise INT8, the robustness of FP6 over INT6 except at very fine granularity, and the competitiveness of INT4 relative to FP4 for most tensors at moderate-to-small block sizes.

\begin{table}[!ht]
\centering
\caption{\textbf{Direct-cast inference  comparisons} across 12 models. RHT denotes random Hadamard rotation. Per-model numbers appear in the Appendix.}\label{tab:inference_summary}
\begin{tabular}{c|cc|cc}
\hline
                  & \multicolumn{2}{c}{Original} & \multicolumn{2}{c}{w/ RHT} \\
                  & INT Win       & FP Win       & INT Win      & FP Win      \\
\hline
MXINT8 v.s. MXFP8 & \textbf{12}   & 0            & \textbf{12}  & 0           \\
MXINT6 v.s. MXFP6 & 0             & \textbf{12}  & 1            & \textbf{11} \\
MXINT4 v.s. MXFP4 & 0             & \textbf{12}  & 0            & \textbf{12} \\
NVINT4 v.s. NVFP4 & 0             & \textbf{12}  & \textbf{12}  & 0   \\
\hline
\end{tabular}
\end{table}

\subsection{Direct-Cast Inference}~\label{sec:inference}

\textbf{Precisions.} For inference, we compare the formats in Table~\ref{tab:low_bit_formats}: MXFP8, MXINT8, MXFP6, MXINT6, MXFP4, MXINT4, NVFP4, and NVINT4. We perform direct-cast inference from a pretrained BFloat16 model and quantize all forward GEMMs.

\textbf{Models.} We evaluate 12 LLMs covering dense and Mixture-of-Experts (MoE) architectures, from 0.6B to 235B parameters: Qwen3-0.6B/1.7B/4B/8B/14B/32B/30B-A3B/235B-A22B~\citep{qwen3}, Llama-3.1-8B/70B, and Llama-3.2-1B/3B~\citep{llama3}. We also apply random Hadamard rotation and quantize $\mathbf{X}\mathbf{R}$ and $\mathbf{R}^\top\mathbf{W}$, where $\mathbf{R}$ is a random Hadamard matrix of size $h \times h$. We set $h$ to the block size (32 for MX formats and 16 for NV formats). We provide official open-source links in Sec.~\ref{sec:reproduction}.

\textbf{Metrics.} Our goal is to compare integer and floating-point low-bit formats under the same settings, so ranking is more informative than absolute accuracy. Following \cite{not_accuracy}, accuracy alone is not sufficient for compressed models because it can hide large behavioral changes. We therefore use distance metrics: specifically, we compute the KL divergence on WikiText2~\citep{wikitext2} between each quantized model and its BFloat16 counterpart. To reduce noise, we compute the divergence over the softmax distribution restricted to the top-25 logits of the BFloat16 model.

\textbf{Results.} Table~\ref{tab:inference_summary} summarizes the comparison between FP and INT formats. Without rotation, MXINT8 outperforms MXFP8 on all 12 models, while MXINT6, MXINT4, and NVINT4 perform worse than MXFP6, MXFP4, and NVFP4. Although NVINT4 and NVFP4 have similar average QSNR in Figure~\ref{fig:qsnr}, NVINT4 loses more often because higher crest factors create more worst-case behavior for integers. With random Hadamard rotation, MXINT8 and NVINT4 win on all 12 models; MXINT6 wins 1 of 12 and MXINT4 loses all 12, consistent with the tensor-wise analysis in Sec.~\ref{sec:tensor}.

\subsection{Training}~\label{sec:training}

% \begin{figure}
%     \centering
%     \includegraphics[width=0.5\linewidth]{figures/loss_curve.pdf}
%     \caption{Loss curves comparison among BF16, MXFP8 and MXINT8 training on Llama-1B with 100B tokens. Results are smoothed by exponential moving average with a coefficient of 0.9.}
%     \label{fig:loss_curve}
% \end{figure}

\begin{wrapfigure}{r}{0.5\textwidth}
    \centering
    % 可选：让图片顶部更贴近当前行
    \vspace{-1.0\baselineskip}
    \includegraphics[width=\linewidth]{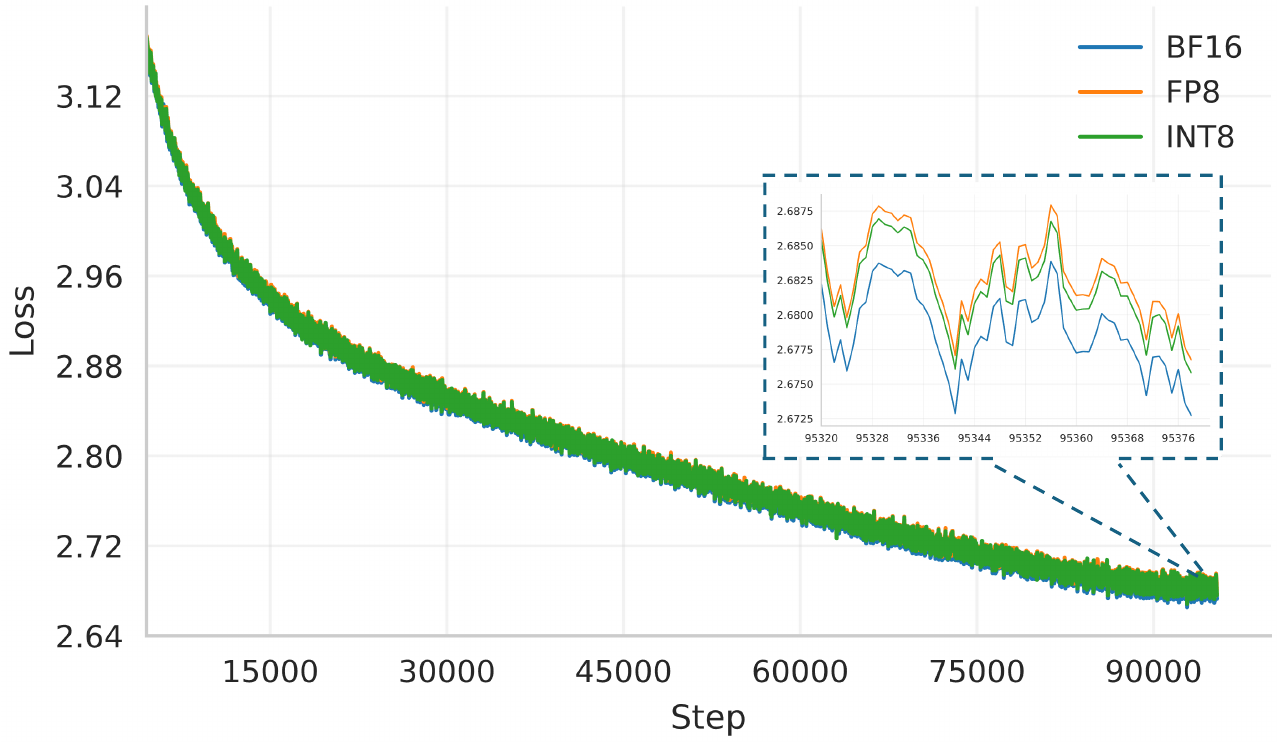}
    \caption{Loss curves comparison among BF16, MXFP8 and MXINT8 training on Llama-1B with 100B tokens. Results are smoothed by exponential moving average with a coefficient of 0.9.}
    \label{fig:loss_curve}
\end{wrapfigure}

\textbf{Precisions.} For training, we focus on nearly lossless low-bit training, which is more practical. Therefore, we study only the 8-bit setting and compare MXINT8 and MXFP8, since FP8 training is demonstrated to be nearly lossless in prior work~\citep{mxfp8_training,liu2024deepseek}.

\textbf{Models and datasets.} We train 1B and 3B Llama3-style~\citep{llama3} models on the OLMo2-Mix-1124~\citep{olmo2} pretraining dataset, with 100B and 200B training tokens, respectively. Detailed model architectures and training hyperparameters are in Sec.~\ref{sec:reproduction}.

\textbf{Metrics.} We measure training performance using two metrics: training loss and task accuracy. We smooth the training loss with an exponential moving average (coefficient $0.9$). We compute all accuracies with \texttt{lm\_eval}~\citep{eval-harness} through 5-shot evaluation. We report \texttt{acc} for WinoGrande~\citep{winogrande} and \texttt{acc\_norm} for HellaSwag~\citep{hellaswag}, Arc\_Challenge, Arc\_Easy~\citep{arc}, PIQA~\citep{piqa}, and Openbookqa~\citep{openbookqa}.

\textbf{Results.} Figure~\ref{fig:loss_curve} shows the loss curves for BF16, MXFP8, and MXINT8 training. The curves for MXFP8 and MXINT8 almost overlap with BF16. In addition, MXINT8 consistently outperforms MXFP8 with a loss that is lower by approximately $0.001$, as shown in the enlarged view in Figure~\ref{fig:loss_curve}. Table~\ref{tab:training} shows that MXINT8 also achieves nearly the same average accuracy across six common-sense reasoning tasks compared to BF16 training. These results demonstrate that MXINT8 supports nearly lossless low-bit training, while existing works~\citep{liu2024deepseek,mxfp8_training} mainly focus on FP8 training.

\begin{table}[!ht]
    \centering
    \setlength{\tabcolsep}{3pt}
    \caption{\textbf{Low-bit training comparisons.} HS, OB, and WG represents Hellaswag, OpenbookQA, and WinoGrande, respectively.}
    \label{tab:training}
    \begin{tabular}{ccccccccccc}
    \hline
    Model size & Training tokens & Precision & loss & Arc\_E & Arc\_C & HS & OB& PIQA & WG & Avg. \\
    \hline
    1B & 100B & BF16 & 2.6727 & 37.80 & 69.40 & 60.20 & 38.40 & 74.43 & 61.09 & 56.89 \\
    \cdashline{1-11}
    1B & 100B & MXFP8 & 2.6767 & 37.03 & 69.82 & 60.28 & 38.00 & 74.37 & 61.64 & 56.86 \\
    1B & 100B & MXINT8 & \textbf{2.6758} & 37.95 & 69.45 & 60.02 & 38.80 & 74.54 & 61.38 & \textbf{57.02} \\
    \hline
    3B & 200B & BF16 & 2.4794 & 46.50 & 75.42 & 72.28 & 45.00 & 78.07 & 69.45 & 64.45 \\
    \cdashline{1-11}
    3B & 200B & MXFP8 & 2.4821 & 46.70 & 74.12 & 72.08 & 44.60 & 77.56 & 69.25 & 64.05 \\
    3B & 200B & MXINT8 & \textbf{2.4812} & 46.10 & 75.58 & 72.00 & 44.80 & 77.78 & 69.55 & \textbf{64.30} \\
    \hline
    \end{tabular}

\end{table}

\begin{table}[!h]
\centering
\caption{Normalized energy and area costs of low-bit formats at same throughput. Single-format results use MXFP8 as the baseline, and mixed-format results use MXFP8+NVFP4 as the baseline.}
\label{tab:energy_area}
\setlength{\tabcolsep}{3pt}
\begin{tabular}{ccccc|cc}
\hline
        & \multicolumn{4}{c}{Single Format} & \multicolumn{2}{c}{Mixed Format} \\
        \hline
        & MXFP8  & MXINT8   & NVFP4 & NVINT4 & MXFP8+NVFP4     & MXINT8+NVINT4     \\
         \hline
Energy & 1x     & \textbf{0.63x}  & 0.55x   & \textbf{0.34x}   & 1x              &    \textbf{0.75x}         \\
\hline
Area   & 1x     & \textbf{0.79x}  & 0.54x   & \textbf{0.38x}   & 1x              &    \textbf{0.66x}         \\
\hline
\end{tabular}
\end{table}
\section{Hardware Cost Analysis}\label{sec:hardware_cost}
Based on the hardware model in Sec.~\ref{sec:hardware_model_cost}, we evaluate the energy and area cost of a Matrix-Multiply Unit (MMU) that supports the MX format.
Table~\ref{tab:energy_area} shows that MXINT8 and NVINT4 reduce energy by 37\% and 38\%, respectively, compared with MXFP8 and NVFP4.
We also evaluate mixed-format configurations. Following the NVIDIA Blackwell GPUs~\citep{nv_quant_doc}, we study a chip that supports both 8-bit and 4-bit data types and set the throughput ratio of 8-bit to 4-bit to 1:2 to match the communication bandwidth. As shown in Table~\ref{tab:energy_area}, the ``MXINT8+NVINT4'' configuration further reduces area by about 34\% relative to ``MXFP8+NVFP4'', mainly because circuit reuse is simpler in the INT pipeline (Table~\ref{tab: reuse_scheme}).
Overall, this analysis shows that, at matched throughput, low-bit integer formats are more hardware-efficient than low-bit floating-point formats.

\section{Conclusion}

% Our comprehensive study reveals a critical crossover point where fine-grained integer (INT) quantization consistently outperforms floating-point (FP) formats for modern LLMs. This finding challenges the current hardware trajectory, as we show INT formats provide a dual advantage of superior accuracy and greater hardware efficiency. We therefore call for a strategic shift in both academia and industry toward algorithm-hardware co-design centered on fine-grained INT to build more powerful and efficient AI accelerators.

Our comprehensive study reveals a critical and nuanced trade-off between integer (INT) and floating-point (FP) quantization. We find that while FP formats are effective at coarse granularities, the popular fine-grained MXINT8 consistently outperforms its FP counterpart MXFP8 in both accuracy and hardware efficiency. For 4-bit formats, the accuracy advantage shifts to FP (MXFP4, NVFP4) , though we demonstrate that NVINT4 can surpass NVFP4 when combined with random Hadamard rotation. These findings challenge the current hardware trajectory, which is increasingly focused on FP. We therefore call for a strategic shift in both academia and industry toward algorithm-hardware co-design that re-evaluates and prioritizes fine-grained INT formats to build more powerful and efficient AI accelerators.

\clearpage

\bibliographystyle{plainnat}
\bibliography{main}

\begin{thebibliography}{45}
\providecommand{\natexlab}[1]{#1}
\providecommand{\url}[1]{\texttt{#1}}
\expandafter\ifx\csname urlstyle\endcsname\relax
  \providecommand{\doi}[1]{doi: #1}\else
  \providecommand{\doi}{doi: \begingroup \urlstyle{rm}\Url}\fi

\bibitem[Ainslie et~al.(2023)Ainslie, Lee-Thorp, De~Jong, Zemlyanskiy, Lebr{\'o}n, and Sanghai]{gqa}
Joshua Ainslie, James Lee-Thorp, Michiel De~Jong, Yury Zemlyanskiy, Federico Lebr{\'o}n, and Sumit Sanghai.
\newblock Gqa: Training generalized multi-query transformer models from multi-head checkpoints.
\newblock \emph{arXiv preprint arXiv:2305.13245}, 2023.

\bibitem[Ashkboos et~al.(2024)Ashkboos, Mohtashami, Croci, Li, Jaggi, Alistarh, Hoefler, and Hensman]{quarot}
Saleh Ashkboos, Amirkeivan Mohtashami, Maximilian~L Croci, Bo~Li, Martin Jaggi, Dan Alistarh, Torsten Hoefler, and James Hensman.
\newblock Quarot: Outlier-free 4-bit inference in rotated llms.
\newblock \emph{arXiv preprint arXiv:2404.00456}, 2024.

\bibitem[Bennett(1948)]{Bennett1948Spectra}
W.~R. Bennett.
\newblock Spectra of quantized signals.
\newblock \emph{Bell System Technical Journal}, 27\penalty0 (3):\penalty0 446--472, July 1948.
\newblock \doi{10.1002/j.1538-7305.1948.tb01364.x}.

\bibitem[Bisk et~al.(2020)Bisk, Zellers, Gao, Choi, et~al.]{piqa}
Yonatan Bisk, Rowan Zellers, Jianfeng Gao, Yejin Choi, et~al.
\newblock Piqa: Reasoning about physical commonsense in natural language.
\newblock In \emph{Proceedings of the AAAI conference on artificial intelligence}, pages 7432--7439, 2020.

\bibitem[Castro et~al.(2025)Castro, Panferov, Tabesh, Sieberling, Chen, Nikdan, Ashkboos, and Alistarh]{quaret}
Roberto~L Castro, Andrei Panferov, Soroush Tabesh, Oliver Sieberling, Jiale Chen, Mahdi Nikdan, Saleh Ashkboos, and Dan Alistarh.
\newblock Quartet: Native fp4 training can be optimal for large language models.
\newblock \emph{arXiv preprint arXiv:2505.14669}, 2025.

\bibitem[Chen et~al.(2024{\natexlab{a}})Chen, Liu, Wang, Bin, Shao, and Luo]{prefixquant}
Mengzhao Chen, Yi~Liu, Jiahao Wang, Yi~Bin, Wenqi Shao, and Ping Luo.
\newblock Prefixquant: Eliminating outliers by prefixed tokens for large language models quantization.
\newblock \emph{arXiv preprint arXiv:2410.05265}, 2024{\natexlab{a}}.

\bibitem[Chen et~al.(2024{\natexlab{b}})Chen, Shao, Xu, Wang, Gao, Zhang, and Luo]{efficientqat}
Mengzhao Chen, Wenqi Shao, Peng Xu, Jiahao Wang, Peng Gao, Kaipeng Zhang, and Ping Luo.
\newblock Efficientqat: Efficient quantization-aware training for large language models.
\newblock \emph{arXiv preprint arXiv:2407.11062}, 2024{\natexlab{b}}.

\bibitem[Chen et~al.(2025{\natexlab{a}})Chen, Zhang, Liu, Zeng, Xue, Liu, Li, Ma, Huang, Zhou, et~al.]{qat_sl}
Mengzhao Chen, Chaoyi Zhang, Jing Liu, Yutao Zeng, Zeyue Xue, Zhiheng Liu, Yunshui Li, Jin Ma, Jie Huang, Xun Zhou, et~al.
\newblock Scaling law for quantization-aware training.
\newblock \emph{arXiv preprint arXiv:2505.14302}, 2025{\natexlab{a}}.

\bibitem[Chen et~al.(2025{\natexlab{b}})Chen, Xi, Zhu, and Chen]{mxfp4_vit}
Yuxiang Chen, Haocheng Xi, Jun Zhu, and Jianfei Chen.
\newblock Oscillation-reduced mxfp4 training for vision transformers.
\newblock \emph{ArXiv}, abs/2502.20853, 2025{\natexlab{b}}.

\bibitem[Clark et~al.(2018)Clark, Cowhey, Etzioni, Khot, Sabharwal, Schoenick, and Tafjord]{arc}
Peter Clark, Isaac Cowhey, Oren Etzioni, Tushar Khot, Ashish Sabharwal, Carissa Schoenick, and Oyvind Tafjord.
\newblock Think you have solved question answering? try arc, the ai2 reasoning challenge.
\newblock \emph{arXiv preprint arXiv:1803.05457}, 2018.

\bibitem[Darvish~Rouhani et~al.(2023)Darvish~Rouhani, Zhao, Elango, Shafipour, Hall, Mesmakhosroshahi, More, Melnick, Golub, Varatkar, et~al.]{qsnr}
Bita Darvish~Rouhani, Ritchie Zhao, Venmugil Elango, Rasoul Shafipour, Mathew Hall, Maral Mesmakhosroshahi, Ankit More, Levi Melnick, Maximilian Golub, Girish Varatkar, et~al.
\newblock With shared microexponents, a little shifting goes a long way.
\newblock In \emph{Proceedings of the 50th Annual International Symposium on Computer Architecture}, pages 1--13, 2023.

\bibitem[Dettmers et~al.(2022)Dettmers, Lewis, Belkada, and Zettlemoyer]{llmint8}
Tim Dettmers, Mike Lewis, Younes Belkada, and Luke Zettlemoyer.
\newblock Gpt3. int8 (): 8-bit matrix multiplication for transformers at scale.
\newblock \emph{Advances in neural information processing systems}, 35:\penalty0 30318--30332, 2022.

\bibitem[Dubey et~al.(2024)Dubey, Jauhri, Pandey, Kadian, Al-Dahle, Letman, Mathur, Schelten, Yang, Fan, et~al.]{llama3}
Abhimanyu Dubey, Abhinav Jauhri, Abhinav Pandey, Abhishek Kadian, Ahmad Al-Dahle, Aiesha Letman, Akhil Mathur, Alan Schelten, Amy Yang, Angela Fan, et~al.
\newblock The llama 3 herd of models.
\newblock \emph{arXiv e-prints}, pages arXiv--2407, 2024.

\bibitem[Dutta et~al.(2024)Dutta, Krishnan, Kwatra, and Ramjee]{not_accuracy}
Abhinav Dutta, Sanjeev Krishnan, Nipun Kwatra, and Ramachandran Ramjee.
\newblock Accuracy is not all you need.
\newblock \emph{Advances in Neural Information Processing Systems}, 37:\penalty0 124347--124390, 2024.

\bibitem[Frantar et~al.(2022)Frantar, Ashkboos, Hoefler, and Alistarh]{gptq}
Elias Frantar, Saleh Ashkboos, Torsten Hoefler, and Dan Alistarh.
\newblock Gptq: Accurate post-training quantization for generative pre-trained transformers.
\newblock \emph{arXiv preprint arXiv:2210.17323}, 2022.

\bibitem[Frantar et~al.(2025)Frantar, Evci, Park, Houlsby, and Alistarh]{compression_sl}
Elias Frantar, Utku Evci, Wonpyo Park, Neil Houlsby, and Dan Alistarh.
\newblock Compression scaling laws: Unifying sparsity and quantization.
\newblock \emph{arXiv preprint arXiv:2502.16440}, 2025.

\bibitem[Gao et~al.(2024)Gao, Tow, Abbasi, Biderman, Black, DiPofi, Foster, Golding, Hsu, Le~Noac'h, Li, McDonell, Muennighoff, Ociepa, Phang, Reynolds, Schoelkopf, Skowron, Sutawika, Tang, Thite, Wang, Wang, and Zou]{eval-harness}
Leo Gao, Jonathan Tow, Baber Abbasi, Stella Biderman, Sid Black, Anthony DiPofi, Charles Foster, Laurence Golding, Jeffrey Hsu, Alain Le~Noac'h, Haonan Li, Kyle McDonell, Niklas Muennighoff, Chris Ociepa, Jason Phang, Laria Reynolds, Hailey Schoelkopf, Aviya Skowron, Lintang Sutawika, Eric Tang, Anish Thite, Ben Wang, Kevin Wang, and Andy Zou.
\newblock A framework for few-shot language model evaluation, 07 2024.
\newblock URL \url{https://zenodo.org/records/12608602}.

\bibitem[Hoffmann et~al.(2022)Hoffmann, Borgeaud, Mensch, Buchatskaya, Cai, Rutherford, Casas, Hendricks, Welbl, Clark, et~al.]{chinchilla}
Jordan Hoffmann, Sebastian Borgeaud, Arthur Mensch, Elena Buchatskaya, Trevor Cai, Eliza Rutherford, Diego de~Las Casas, Lisa~Anne Hendricks, Johannes Welbl, Aidan Clark, et~al.
\newblock Training compute-optimal large language models.
\newblock \emph{arXiv preprint arXiv:2203.15556}, 2022.

\bibitem[Kumar et~al.(2024)Kumar, Ankner, Spector, Bordelon, Muennighoff, Paul, Pehlevan, R{\'e}, and Raghunathan]{sl_precision_harvard}
Tanishq Kumar, Zachary Ankner, Benjamin~F Spector, Blake Bordelon, Niklas Muennighoff, Mansheej Paul, Cengiz Pehlevan, Christopher R{\'e}, and Aditi Raghunathan.
\newblock Scaling laws for precision.
\newblock \emph{arXiv preprint arXiv:2411.04330}, 2024.

\bibitem[Lin et~al.(2023)Lin, Tang, Tang, Yang, Dang, and Han]{awq}
Ji~Lin, Jiaming Tang, Haotian Tang, Shang Yang, Xingyu Dang, and Song Han.
\newblock Awq: Activation-aware weight quantization for llm compression and acceleration.
\newblock \emph{arXiv preprint arXiv:2306.00978}, 2023.

\bibitem[Liu et~al.(2024{\natexlab{a}})Liu, Feng, Xue, Wang, Wu, Lu, Zhao, Deng, Zhang, Ruan, et~al.]{liu2024deepseek}
Aixin Liu, Bei Feng, Bing Xue, Bingxuan Wang, Bochao Wu, Chengda Lu, Chenggang Zhao, Chengqi Deng, Chenyu Zhang, Chong Ruan, et~al.
\newblock Deepseek-v3 technical report.
\newblock \emph{arXiv preprint arXiv:2412.19437}, 2024{\natexlab{a}}.

\bibitem[Liu et~al.(2024{\natexlab{b}})Liu, Zhao, Fedorov, Soran, Choudhary, Krishnamoorthi, Chandra, Tian, and Blankevoort]{spinquant}
Zechun Liu, Changsheng Zhao, Igor Fedorov, Bilge Soran, Dhruv Choudhary, Raghuraman Krishnamoorthi, Vikas Chandra, Yuandong Tian, and Tijmen Blankevoort.
\newblock Spinquant: Llm quantization with learned rotations.
\newblock \emph{arXiv preprint arXiv:2405.16406}, 2024{\natexlab{b}}.

\bibitem[Liu et~al.(2025)Liu, Zhao, Huang, Chen, Zhang, Zhao, Roy, Jin, Xiong, Shi, et~al.]{paretoq}
Zechun Liu, Changsheng Zhao, Hanxian Huang, Sijia Chen, Jing Zhang, Jiawei Zhao, Scott Roy, Lisa Jin, Yunyang Xiong, Yangyang Shi, et~al.
\newblock Paretoq: Scaling laws in extremely low-bit llm quantization.
\newblock \emph{arXiv preprint arXiv:2502.02631}, 2025.

\bibitem[Markstein(2008)]{ieee_float}
Peter Markstein.
\newblock The new ieee-754 standard for floating point arithmetic.
\newblock Schloss Dagstuhl--Leibniz-Zentrum f{\"u}r Informatik, 2008.

\bibitem[Merity et~al.(2016)Merity, Xiong, Bradbury, and Socher]{wikitext2}
Stephen Merity, Caiming Xiong, James Bradbury, and Richard Socher.
\newblock Pointer sentinel mixture models.
\newblock \emph{arXiv preprint arXiv:1609.07843}, 2016.

\bibitem[Mihaylov et~al.(2018)Mihaylov, Clark, Khot, and Sabharwal]{openbookqa}
Todor Mihaylov, Peter Clark, Tushar Khot, and Ashish Sabharwal.
\newblock Can a suit of armor conduct electricity? a new dataset for open book question answering.
\newblock \emph{arXiv preprint arXiv:1809.02789}, 2018.

\bibitem[Mishra et~al.(2025)Mishra, Stosic, and Layton]{mxfp8_training}
Asit Mishra, Dusan Stosic, and Simon Layton.
\newblock Recipes for pre-training llms with mxfp8.
\newblock \emph{arXiv preprint arXiv:2506.08027}, 2025.

\bibitem[Norrie et~al.(2021)Norrie, Patil, Yoon, Kurian, Li, Laudon, Young, Jouppi, and Patterson]{tpu}
Thomas Norrie, Nishant Patil, Doe~Hyun Yoon, George Kurian, Sheng Li, James Laudon, Cliff Young, Norman Jouppi, and David Patterson.
\newblock The design process for google's training chips: Tpuv2 and tpuv3.
\newblock \emph{IEEE Micro}, 41\penalty0 (2):\penalty0 56--63, 2021.
\newblock \doi{10.1109/MM.2021.3058217}.

\bibitem[{NVIDIA Corporation}(2020)]{NVIDIA_A100}
{NVIDIA Corporation}.
\newblock Nvidia a100 tensor core gpu architecture.
\newblock Whitepaper, NVIDIA Corporation, 2020.
\newblock URL \url{https://www.nvidia.com/en-us/data-center/ampere-architecture/}.

\bibitem[{NVIDIA Corporation}(2022)]{NVIDIA_H100}
{NVIDIA Corporation}.
\newblock Nvidia h100 tensor core gpu architecture.
\newblock Whitepaper, NVIDIA Corporation, 2022.
\newblock URL \url{https://www.nvidia.com/en-us/data-center/hopper-architecture/}.

\bibitem[{NVIDIA Corporation}(2024{\natexlab{a}})]{NVIDIA_Blackwell}
{NVIDIA Corporation}.
\newblock Nvidia blackwell gpu architecture.
\newblock Whitepaper, NVIDIA Corporation, 2024{\natexlab{a}}.
\newblock URL \url{https://www.nvidia.com/en-us/data-center/blackwell-architecture/}.

\bibitem[{NVIDIA Corporation}(2024{\natexlab{b}})]{nv_quant_doc}
{NVIDIA Corporation}.
\newblock Working with quantized types -- nvidia tensorrt documentation.
\newblock \url{https://docs.nvidia.com/deeplearning/tensorrt/latest/inference-library/work-quantized-types.html}, 2024{\natexlab{b}}.
\newblock Accessed: 2025-09-03.

\bibitem[OLMo et~al.(2024)OLMo, Walsh, Soldaini, Groeneveld, Lo, Arora, Bhagia, Gu, Huang, Jordan, et~al.]{olmo2}
Team OLMo, Pete Walsh, Luca Soldaini, Dirk Groeneveld, Kyle Lo, Shane Arora, Akshita Bhagia, Yuling Gu, Shengyi Huang, Matt Jordan, et~al.
\newblock 2 olmo 2 furious.
\newblock \emph{arXiv preprint arXiv:2501.00656}, 2024.

\bibitem[Rouhani et~al.(2023)Rouhani, Zhao, More, Hall, Khodamoradi, Deng, Choudhary, Cornea, Dellinger, Denolf, et~al.]{ocp}
Bita~Darvish Rouhani, Ritchie Zhao, Ankit More, Mathew Hall, Alireza Khodamoradi, Summer Deng, Dhruv Choudhary, Marius Cornea, Eric Dellinger, Kristof Denolf, et~al.
\newblock Microscaling data formats for deep learning.
\newblock \emph{arXiv preprint arXiv:2310.10537}, 2023.

\bibitem[Sakaguchi et~al.(2021)Sakaguchi, Bras, Bhagavatula, and Choi]{winogrande}
Keisuke Sakaguchi, Ronan~Le Bras, Chandra Bhagavatula, and Yejin Choi.
\newblock Winogrande: An adversarial winograd schema challenge at scale.
\newblock \emph{Communications of the ACM}, 64\penalty0 (9):\penalty0 99--106, 2021.

\bibitem[Shao et~al.(2023)Shao, Chen, Zhang, Xu, Zhao, Li, Zhang, Gao, Qiao, and Luo]{omniquant}
Wenqi Shao, Mengzhao Chen, Zhaoyang Zhang, Peng Xu, Lirui Zhao, Zhiqian Li, Kaipeng Zhang, Peng Gao, Yu~Qiao, and Ping Luo.
\newblock Omniquant: Omnidirectionally calibrated quantization for large language models.
\newblock \emph{arXiv preprint arXiv:2308.13137}, 2023.

\bibitem[Shazeer(2020)]{swiglu}
Noam Shazeer.
\newblock Glu variants improve transformer.
\newblock \emph{arXiv preprint arXiv:2002.05202}, 2020.

\bibitem[Sun et~al.(2024)Sun, Chen, Kolter, and Liu]{massive}
Mingjie Sun, Xinlei Chen, J~Zico Kolter, and Zhuang Liu.
\newblock Massive activations in large language models.
\newblock \emph{arXiv preprint arXiv:2402.17762}, 2024.

\bibitem[Tseng et~al.(2025)Tseng, Yu, and Park]{mxfp4_training}
Albert Tseng, Tao Yu, and Youngsuk Park.
\newblock Training llms with mxfp4.
\newblock \emph{arXiv preprint arXiv:2502.20586}, 2025.

\bibitem[Ul~Haq et~al.(2025)Ul~Haq, El-Maleh, and Alsuwaiyan]{fp_adders}
Sami Ul~Haq, Aiman~H. El-Maleh, and Ali Alsuwaiyan.
\newblock Multiple-input floating-point adders: A comprehensive review.
\newblock \emph{IEEE Access}, 13:\penalty0 91012--91024, 2025.
\newblock \doi{10.1109/ACCESS.2025.3572430}.

\bibitem[Xiao et~al.(2023)Xiao, Lin, Seznec, Wu, Demouth, and Han]{smoothquant}
Guangxuan Xiao, Ji~Lin, Mickael Seznec, Hao Wu, Julien Demouth, and Song Han.
\newblock Smoothquant: Accurate and efficient post-training quantization for large language models.
\newblock In \emph{International Conference on Machine Learning}, pages 38087--38099. PMLR, 2023.

\bibitem[Yang et~al.(2025)Yang, Li, Yang, Zhang, Hui, Zheng, Yu, Gao, Huang, Lv, et~al.]{qwen3}
An~Yang, Anfeng Li, Baosong Yang, Beichen Zhang, Binyuan Hui, Bo~Zheng, Bowen Yu, Chang Gao, Chengen Huang, Chenxu Lv, et~al.
\newblock Qwen3 technical report.
\newblock \emph{arXiv preprint arXiv:2505.09388}, 2025.

\bibitem[Yuan et~al.(2024)Yuan, Shang, Zhou, Dong, Zhou, Xue, Wu, Li, Gu, Lee, et~al.]{unveiled}
Zhihang Yuan, Yuzhang Shang, Yang Zhou, Zhen Dong, Zhe Zhou, Chenhao Xue, Bingzhe Wu, Zhikai Li, Qingyi Gu, Yong~Jae Lee, et~al.
\newblock Llm inference unveiled: Survey and roofline model insights.
\newblock \emph{arXiv preprint arXiv:2402.16363}, 2024.

\bibitem[Zellers et~al.(2019)Zellers, Holtzman, Bisk, Farhadi, and Choi]{hellaswag}
Rowan Zellers, Ari Holtzman, Yonatan Bisk, Ali Farhadi, and Yejin Choi.
\newblock Hellaswag: Can a machine really finish your sentence?
\newblock \emph{arXiv preprint arXiv:1905.07830}, 2019.

\bibitem[Zhang et~al.(2024)Zhang, Zhao, Cao, Zhang, Wang, Cao, Yang, Yang, Zhang, and Xu]{fp_vs_int}
Yijia Zhang, Lingran Zhao, Shijie Cao, Sicheng Zhang, Wenqiang Wang, Ting Cao, Fan Yang, Mao Yang, Shanghang Zhang, and Ningyi Xu.
\newblock Integer or floating point? new outlooks for low-bit quantization on large language models.
\newblock In \emph{2024 IEEE International Conference on Multimedia and Expo (ICME)}, pages 1--6. IEEE, 2024.

\end{thebibliography}

\clearpage

\beginappendix

% \newpage
% \appendix
% \section{Appendix}

\section*{Outlines}
\begin{itemize}
\item Sec.~\ref{sec:related_works} introduces related works.
\item 
Sec.~\ref{sec:proof_of_theorem} details the proofs of Theorems~1 and~2 on INT and FP QSNR estimation.
\item Sec.~\ref{sec:hardware_model_cost} presents the hardware cost estimation model.
\item Sec.~\ref{sec:reproduction} provides additional details on the models used and ablation studies, and reports the numerical results corresponding to the figures in the main paper.
\end{itemize}
% \section*{Usage of Large Language Models}
% We use LLMs to polish the paper, correct the grammar, and for some of the figures in the article, the initial drawing codes are generated by LLMs.

\section{Related Work}\label{sec:related_works}
\textbf{Quantization Algorithms.} Quantization methods include post-training quantization (PTQ)~\citep{awq,gptq,omniquant,smoothquant} and quantization-aware training (QAT)~\citep{efficientqat,paretoq}, which speed up inference. Low-bit training~\citep{mxfp8_training,mxfp4_training,mxfp4_vit} speeds up both training and inference. Several works also study scaling laws~\citep{chinchilla} for low-bit quantization~\citep{quaret,qat_sl,compression_sl,sl_precision_harvard}. However, most prior work focuses on a single low-bit format—either integer or floating-point—and does not provide direct comparisons between these formats. \cite{fp_vs_int} study mixed-format quantization in the PTQ setting, assigning integer or floating-point formats to different model parts.

\textbf{Hardware.} Previous accelerators~\citep{NVIDIA_A100,NVIDIA_H100} do not natively support fine-grained quantization, so algorithms~\citep{smoothquant,prefixquant} face challenges with per-channel quantization in the presence of outliers~\citep{massive}. Recently, OCP~\cite{ocp} proposes Microscaling (MX) data formats, which combine a per-block scaling factor with a block size of 32 to improve low-bit quantization performance. NVIDIA Blackwell~\citep{NVIDIA_Blackwell} supports MXFP8, MXFP4, and NVFP4 at the hardware level.

% In this paper, we offer a comprehensive comparison of integer and float-point quantization, demonstrate that integer can consistently outperforms float-point quantization in fine-grained quantization, as well as lower hardware cost. 
% %

% \textbf{LLM Quantization}

% \textbf{LLM low-bits training}

% \textbf{AI Hardware Architectures}

% % 明确指出本文与已有工作的不同之处——我们提供了首个跨越不同粒度的INT与FP的系统性对比，并连接了算法与硬件两个领域。
% Low-bits quantization formats determines the representation

\section{Proofs of Theorems}\label{sec:proof_of_theorem}

\subsection{Common assumptions and notation}
\label{sec:appendix_common_assumptions}

We consider block vectors $\mathbf{X}\in\mathbb{R}^g$ with i.i.d. entries $X_i\sim\mathcal{N}(0,\sigma^2)$. We denote the block RMS by $\sigma := \mathrm{RMS}(\mathbf{X})$ and the crest factor by
\begin{equation}~\label{eq:appen_kappa}
\kappa \;:=\; \frac{\max(|\mathbf{X}|)}{\sigma}.
\end{equation}

For MX format, which uses blockwise UE8M0 scale factors, we set
\begin{equation}~\label{eq:appen_s'}
s' \;=\; 2^{\lceil \log_2 s \rceil} \;=\; \rho\,s,\qquad \rho\in[1,2),
\end{equation}
and choose $s'\ge s$ to avoid upper clipping. When the scale factors use BFloat16 or E4M3, we set $\rho=1$. The ideal scale $s$ matches the largest codebook magnitude to the block maximum:
\begin{equation}~\label{eq:appen_s}
s \;=\; \frac{\max(|\mathbf{X}|)}{Q_{\mathrm{ref}}},
\end{equation}
where $Q_{\mathrm{ref}}$ depends on the target format:
\begin{itemize}
\item INT$(b)$ (symmetric): $Q_{\mathrm{ref}} = Q := 2^{b-1}-1$ (largest integer code).
\item FP$(E,M,B)$ (with subnormals): $Q_{\mathrm{ref}} = Q_{\max}$ (largest finite normal magnitude; e.g., $Q_{\max}=448$ for E4M3).
\end{itemize}
This convention matches the main text: we reuse $(\sigma,\kappa,\rho,s,s')$, and $s'\ge s$ prevents overflow for both INT and FP quantization. Unless stated otherwise, expectations are over both the data and the quantization randomness, and $\|\mathbf{X}\|^2 \approx k\sigma^2$.

\subsection{Theorem 1 (INT quantization)}
\label{sec:theorem1_proof}

\textbf{INT quantization.}
We consider a symmetric, uniform quantizer with bit-width $b$ and integer range $[-Q,Q]$, where
\begin{equation}~\label{eq:appen_q}
Q \;=\; 2^{b-1}-1 \quad\text{(e.g., $Q\in\{127,31,7\}$ for $b\in\{8,6,4\}$)}.
\end{equation}
The quantize--dequantize operation is
\begin{equation}
\mathbf{X}_q \;=\; \operatorname{clamp}\!\big(\operatorname{round}(\tfrac{\mathbf{X}}{s'}),\, -Q,\, Q \big)\cdot s',
\end{equation}
so the effective step in the quantization is $\Delta := s'$.

\textbf{Error model.}
Let the elementwise error be $e := {X}-{X}_q$. For a non-saturating symmetric quantizer with round-to-nearest, $e\in[-\frac{\Delta}{2},\,\frac{\Delta}{2}]$. Under the standard high-resolution model~\cite{Bennett1948Spectra}, the error is approximately uniform and independent of $\mathbf{X}$:
\begin{equation}
\mathbb{E}[e]=0,
\qquad
\mathbb{E}[e^2]=\frac{\Delta^2}{12}.
\end{equation}

\textbf{QSNR.}
Define the QSNR as
\begin{equation}
\mathrm{QSNR}
\;=\;
-10\log_{10}\!\left(\frac{\|\mathbf{X}-\mathbf{X}_q\|^2}{\|\mathbf{X}\|^2}\right).
\end{equation}
We have $\mathbb{E}[\|\mathbf{X}\|^2]\approx k\sigma^2$ and
$\mathbb{E}[\|\mathbf{X}-\mathbf{X}_q\|^2]\approx k\,\mathbb{E}[e^2]=k\Delta^2/12$, hence
\begin{equation}
\mathrm{QSNR}
\;\approx\;
-10\log_{10}\!\left(\frac{\Delta^2}{12\,\sigma^2}\right).
\end{equation}

\textbf{Expressing $\Delta$ via crest factor and scale overhead.}
Using Eq.~(\ref{eq:appen_kappa}--\ref{eq:appen_s}), 
\begin{equation}
\Delta \;=\; s' \;=\; \frac{\rho\,\kappa\,\sigma}{Q}.
\end{equation}
Substituting into the QSNR expression gives
\begin{equation}
\frac{\Delta^2}{12\,\sigma^2}
\;=\;
\frac{(\rho\,\kappa)^2}{12\,Q^2},
\end{equation}
and therefore
\begin{equation}
\boxed{\;
\mathrm{QSNR_{MXINT}}
\;\approx\;
-10\log_{10}\!\left(\frac{\kappa^2}{12\,Q^2}\right)
\;\approx\;
4.78
\;+\; 6.02\,b
\;-\; 20\log_{10}(\rho)
\;-\; 20\log_{10}(\kappa)
\;}
\end{equation}
where we use $Q \approx 2^{b-1}$ in Eq.~(\ref{eq:appen_q}). This form makes explicit: (i) $\approx 6.02$ dB per additional bit, (ii) up to $6.02$ dB loss from the power-of-two overhead ($\rho\in[1,2)$), and (iii) a penalty that scales with the crest factor $\kappa$ (which typically increases with larger block size).

\textbf{Extension to high-precision scale factors.}
The analysis above assumes UE8M0 scaling, which rounds the scale and introduces the overhead $\rho\in[1,2)$. With the E4M3 scale format used in NVINT4, the per-block scale closely matches the ideal value, so $\rho \approx 1$, and the element at the block maximum maps with (near-)zero error. For block size $g$ (elements per block), the INT QSNR with an E4M3 scale is
\begin{equation}
\boxed{\;
\mathrm{QSNR_{NVINT}}
\;\approx\;
-10\log_{10}\!\left(\frac{\kappa^2}{12\,Q^2}\cdot \frac{g-1}{g}\right)
\;=\;
4.78
\;+\; 6.02\,b
\;-\; 20\log_{10}(\kappa)
\;+\; 10\log_{10}\!\left(\frac{g}{g-1}\right)
\;}
\label{eq:int_qsnr_e4m3}
\end{equation}
where $10\log_{10}\!\big(\tfrac{g}{g-1}\big)$ accounts for one (near) error-free element per block.

\subsection{Theorem 2 (FP quantization)}
\label{sec:theorem2_proof}

\textbf{FP quantization.}
Consider a target floating-point format FP$(E,M,B)$ with sign, $E$ exponent bits (bias $B$), and $M$ mantissa bits, with subnormals enabled. The representable numbers split into normal and subnormal domains:
\begin{equation}
\mathbb{C}_\text{FP} = \begin{cases}
(-1)^{s} \times (1.m)_2 \times 2^{e-\text{bias}} & \text{if } e \neq 0 \text{ (Normal)}, \\
(-1)^{s} \times (0.m)_2 \times 2^{1-\text{bias}} & \text{if } e = 0,\,m \neq 0 \text{ (Subnormal)},
\end{cases}
\end{equation}
where $s$, $e$, and $m$ are the sign, exponent, and mantissa of a floating-point number. Let $Q_{\max}$ denote the largest finite normal magnitude (e.g., $Q_{\max}=448$ for E4M3), and let $N_{\min} := 2^{1-B}$ be the smallest normal. We also define the subnormal spacing in the codebook as $S_{\min} := 2^{1-B-M}$.

We use a block scale $s'$ (Eq.(\ref{eq:appen_s'})) and perform quantize--dequantize as
\begin{equation}
\mathbf{X}_q \;=\; s'\cdot \operatorname{Nearest}\!\Big(\tfrac{\mathbf{X}}{s'},\,\mathbb{C}_{\mathrm{FP}}\Big),
\end{equation}
where $\mathbb{C}_{\mathrm{FP}}$ is the FP codebook. We choose the ideal scale $s = \max(|\mathbf{X}|)/Q_{\max}$ and set $s'=\rho s$ with $\rho\in[1,2)$ for UE8M0 (power-of-two) scaling; $\rho\approx 1$ when the scale uses E4M3.

\textbf{Error decomposition.}
Let $e := \mathbf{X}-\mathbf{X}_q$. We study the relative MSE
\begin{equation}
R \;:=\; \frac{\mathbb{E}[e^2]}{\mathbb{E}[\mathbf{X}^2]} \;=\; \frac{\mathbb{E}[e^2]}{\sigma^2},
\qquad
\mathrm{QSNR} \;:=\; -10\log_{10}R.
\end{equation}
Under a high-resolution model~\cite{Bennett1948Spectra}, the within-cell error is unbiased and uniform on $[-\frac{\Delta}{2},\frac{\Delta}{2}]$, and the logarithmic phase
\begin{equation}
r \;:=\; 2^{\{\log_2(|X|/s')\}} \in [1,2)
\end{equation}
(the fractional part $\{\cdot\}$ of $\log_2(|X|/s')$) is approximately uniform on $[1,2)$.

Define the signal-domain normal threshold $T_N$ and the subnormal step $\Delta_{\mathrm{sub}}$ as
\begin{equation}
T_N := s' N_{\min},
\qquad
\Delta_{\mathrm{sub}} := s'\,S_{\min} = s'\,2^{1-B-M}.
\end{equation}
We split the amplitude axis into normal and subnormal regions:

\begin{itemize}
\item \textbf{Normal region ($|X|\ge T_N$).}
Let $e(X):=\lfloor \log_2(\tfrac{|X|}{s'})\rfloor$ be the exponent bin of $\tfrac{X}{s'}$.
The local effective quantization step is
\begin{equation}
\Delta(X) \;=\; s'\,2^{\,e(X)-M}.
\end{equation}
Writing $2^{e(X)} = \tfrac{|X|}{s'r}$ with $r\in[1,2)$ gives
\begin{equation}
\Delta(X) \;=\; \frac{|X|}{r}\,2^{-M}.
\end{equation}
Uniform-error modeling yields
$\mathbb{E}[e^2 \mid X, |X|\ge T_N] = \tfrac{\Delta(X)^2}{12} = \tfrac{|X|^2\,2^{-2M}}{12\,r^2}$.
Averaging over $r\sim\mathrm{Uniform}[1,2]$ gives $\mathbb{E}[1/r^2]=\int_1^2 r^{-2}\,dr = 1/2$, hence
\begin{equation}
\mathbb{E}[e^2 \cdot \mathbf{1}\{|X|\ge T_N\}]
\;\approx\; \alpha_M \,\mathbb{E}[X^2 \cdot \mathbf{1}\{|X|\ge T_N\}],
\quad
\alpha_M := \frac{1}{24\cdot 2^{2M}}.
\end{equation}

\item \textbf{Subnormal but nonzero region ($|X| < T_N$).}
Here the absolute spacing is constant, $\Delta_{\mathrm{sub}}$, so
\begin{equation}
\mathbb{E}[e^2 \mid  |X| < T_N]
\;\approx\; \frac{\Delta_{\mathrm{sub}}^2}{12}
\;=\; \frac{s'^2\,2^{2(1-B-M)}}{12}.
\end{equation}
Let $p_{\mathrm{sub}} := \mathbb{P}(|X| < T_N)$.
Then
\begin{equation}
\mathbb{E}[e^2 \cdot \mathbf{1}\{ |X| < T_N\}]
\;\approx\; \frac{s'^2\,2^{2(1-B-M)}}{12}\,p_{\mathrm{sub}}.
\end{equation}
\end{itemize}

Summing the two contributions and normalizing by $\sigma^2$ yields
\begin{equation}
\frac{\mathbb{E}[e^2]}{\sigma^2}
\;\approx\;
\alpha_M \, w_{\mathrm{norm}}
\;+\;
\beta \,(\rho\,\kappa)^2 \, p_{\mathrm{sub}},
\end{equation}
where we define the dimensionless weight
\begin{equation}
w_{\mathrm{norm}} := \frac{\mathbb{E}[X^2 \cdot \mathbf{1}\{|X|\ge T_N\}]}{\sigma^2},
\end{equation}
and use $\tfrac{s'^2}{\sigma^2} = \tfrac{(\rho\kappa)^2}{Q_{\max}^2}$ with
\begin{equation}
\beta := \frac{2^{2(1-B-M)}}{12\,Q_{\max}^2}.
\end{equation}
Therefore,
\begin{equation}\label{eq:fp_qsnr_appdix}
\boxed{\;
\mathrm{QSNR_{MXFP}} \;\approx\; 
-10\log_{10} \!\big(\alpha_M \, w_{\mathrm{norm}}
\;+\;
\beta \,(\rho\,\kappa)^2 \, p_{\mathrm{sub}}
\big)
\;}
\end{equation}

In the ample dynamic-range regime ($w_{\mathrm{norm}}\approx 1$ and $p_{\mathrm{sub}}\approx 0$),
the law simplifies to
\begin{equation}
\mathrm{QSNR}
\;\approx\;
-10\log_{10}(\alpha_M)
\;=\; 13.80 \text{ dB} \;+\; 6.02\,M \text{ dB},
\end{equation}
independent of block granularity and the distribution of $\mathbf{X}$.

\textbf{Extension to high-precision scale factors.}
The analysis above assumes a UE8M0-quantized scale, which forces $s'$ to be a power of two and introduces the overhead $\rho\in[1,2)$. When the per-block scale uses E4M3 (as in NVFP4), the scale closely tracks the ideal value, so $\rho\approx 1$, and the element at the block maximum maps with negligible error (its scaled value hits $Q_{\max}$). It is therefore natural to exclude the block-maximum contribution from the normal-region error budget. Let $g$ be the block size and define the energy fraction of the block maximum as
\begin{equation}\label{eq:max_energy}
    \eta \;:=\; \frac{\max(|\mathbf{X}|)^2}{g\,\sigma^2} \;=\; \frac{\kappa^2}{g}.
\end{equation}
Setting $\rho=1$ and replacing $w_{\mathrm{norm}}$ by $w_{\mathrm{norm}}-\eta$ in Eq.~\eqref{eq:fp_qsnr_appdix} yields the refined QSNR approximation for FP quantization with an E4M3 scale:
\begin{equation}\label{eq:fp_qsnr_e4m3}
\boxed{\;
\mathrm{QSNR_{NVFP}} \;\approx\; 
-10\log_{10} \!\big(\alpha_M \, (w_{\mathrm{norm}}-\tfrac{\kappa^2}{g})
\;+\;
\beta \,\kappa^2 \, p_{\mathrm{sub}}
\big)
\;}
\end{equation}
This adjustment isolates the block maximum and tightens the prediction when the scale is represented with sufficient precision.

\begin{table}[h]
\centering
\caption{Gate-complexity model for the MAC Unit with $k$ lanes. Here $x$ and $y$ denote exponent and mantissa widths; for INT, $x{=}0$. The aligner width $n$ is given by~\eqref{eq: aligner_width}. ``Main Cells'' list dominant standard cells used in aggregation.}
\label{tab: mac_complexity}
\begin{tabular}{lccccc}
\toprule
Sub-block              & INT Mul                  & FP Mul                   & INT Add            & FP Add              & Main Cells \\ 
\midrule
Multiplier             & $k(x{+}y{+}1)^2$         & $k(y{+}1)^2$             & --                 & --                  & AND, FA, HA \\
Adder (mantissa/int)   & --                       & --                       & $2k(x{+}y{+}1)$     & $kn$                & FA, HA \\
Exponent adder         & --                       & $kx$                     & --                 & --                  & FA, HA \\
Exponent subtractor    & --                       & --                       & --                 & $kx$                & XOR, FA, HA \\
Comparator             & --                       & --                       & --                 & $kx$                & XOR, AND, OR \\
Aligner (barrel)       & --                       & --                       & --                 & $k\,n\log_2 n$      & MUX \\
Normalizer (shared)    & --                       & --                       & --                 & $n\log_2 n$         & MUX, OR \\
\bottomrule
\end{tabular}
\end{table}

\begin{table}[]
\centering
\begin{tabular}{cc}
\hline
Throughput Ratio   & INT8 : INT4 = 1 : 2                   \\ \hline
No reuse           & 1 * int8\_MAC\_unit + 2 * int4\_MAC\_unit \\
INT reuse scheme 1 & 1 * int8\_MAC\_unit + 1 * int4\_MAC\_unit \\
INT reuse scheme 2 & 2 * int8\_(u)int4\_MAC\_unit                 \\ \hline
Throughput Ratio   & FP8 : FP4 = 1 : 2                     \\ \hline
No reuse           & 1 * e4m3\_MAC\_unit + 2 * e2m1\_MAC\_unit \\
FP reuse scheme    & 1 * e4m3\_MAC\_unit + 1 * e2m1\_MAC\_unit \\ \hline
\end{tabular}
\caption{Comparison of MAC unit configurations with the same lanes for different reuse schemes. Notes: (1) No reuse: Highest energy efficiency for INT8 and INT4, but greatest area wastage; (2) INT reuse scheme 1: Use int8 lane as an int4 path directly (set the 8-b input to XXXX\_0000), a little more energy cost for INT4 but lower area cost; (3) INT reuse scheme 2: Use two int8$\times$(u)int4 lanes to reconfigure int8 lane or int4 lane, a little more energy cost for both INT4 and INT8, but lowest area cost; (4) No reuse: Highest energy efficiency for FP8 and FP4, but greatest area wastage; (5) FP reuse scheme: Use fp8 lane as an fp4 path directly (set the 8-b input to S\_00XX\_X00), a little more energy cost for FP4 but lower area cost. We adopt INT reuse scheme 2 and FP reuse scheme to evaluate the area cost shown in Table~\ref{tab:energy_area}.}
\label{tab: reuse_scheme}
\end{table}

\section{Hardware Cost Modeling}~\label{sec:hardware_model_cost}
% \begin{figure}[!h]
%   \centering
%   \includegraphics[width=0.7\linewidth]{figures/hardware/microarchitecture diagram.pdf}
%   \caption{Microarchitecture diagram of hardware implementation.}
%   \label{fig: microarchitecture diagram}
% \end{figure}
\textbf{Scope and assumptions.}
We develop a compact gate-level model to estimate the chip area and energy of a GEMM engine under low-precision formats. Specifically, a low-bit GEMM engine uses four components: a quantizer, a multiply-and-accumulate (MAC) unit, a dequantizer, and an FP32 accumulator. The proposed model accounts only for the MAC unit, a shared FP32 accumulator and a dequantizer; the quantizer is excluded from all cost accounting. In MX/NV formats, the VPU implements quantization by shift/divide-and-round, and the accumulation pipeline can fuse dequantization as two 8-bit integer additions for UE8M0 scale or two floating-point multiplications for E4M3 scale. We omit the quantizer block in VPU to isolate the cost driven by multiplication and accumulation. Unless otherwise stated, we take cell factors from a TSMC FinFET standard-cell library. We model only combinational logic; we ignore sequential elements, placement and routing, and interconnect to enable technology-aware, relative comparisons.

\textbf{Design choice: FP32 accumulation and MMU integration.}
A high-throughput Matrix-Multiply Unit (MMU), as in TPU-like designs~\citep{tpu}, integrates the multiply-and-accumulate datapath and downstream accumulation to improve performance and energy efficiency. To prevent error growth and preserve scalability, we accumulate in FP32. Under the same nominal bit width, FP multipliers are typically more area- and energy-efficient than INT multipliers, whereas FP adders are more expensive than INT adders due to exponent comparison/subtraction, mantissa alignment, and normalization~\citep{fp_vs_int}. With a uniform-alignment design~\cite{fp_adders}, the normalizer count reduces to one shared instance across the $k$ MAC lanes, and we divide its cost by $k$.

\textbf{Mantissa aligner width.}
The mantissa aligner couples accuracy and cost: its bit width $n$ affects numerical fidelity and hardware complexity. We set
\begin{equation}
\label{eq: aligner_width}
	n \;=\; \min\!\bigl(2^{x+1} + 2y,\; \texttt{psum\_bit\_width}\bigr),
\end{equation}
where $x$ and $y$ denote exponent and mantissa widths, respectively (for INT formats, $x\!=\!0$). In all evaluations we use $k\!=\!32$ for MX formats and $k\!=\!16$ for NV formats, and $\texttt{psum\_bit\_width}\!=\!24$.

\textbf{MAC unit structure and sub-blocks.}
We model the MAC unit as a $k$-lane array. Each lane comprises one multiplier. The adders from all lanes are fused together to form a multi-input adder tree structure, incorporating FP-specific alignment and normalization logic. Table~\ref{tab: mac_complexity} reports the dominant logic count (up to constant factors) for the main sub-blocks, where ``Main Cells'' indicate the standard-cell types used for area/energy aggregation. For FP multiplication, we multiply only mantissas and include an exponent adder. For FP addition, we model exponent comparator/subtractor, a barrel aligner, a wide mantissa adder, and one shared normalizer. For INT, we set $x\!=\!0$ in the expressions.

\textbf{Area and energy aggregation for MAC.}
Let $\mathcal{S}$=\{Multiplier, Adder(mantissa/int), Exponent adder, Exponent subtractor, Comparator, Aligner(barrel), Normalizer(shared)\} be the set of sub-block types,
and $\mathcal{G}=\{\text{FA},\text{HA},\text{XOR},\text{AND},\text{OR},\text{MUX}\}$ be the set of cell types with technology-dependent area and energy factors $A_g$ and $E_g$ obtained from the standard-cell library. Let $\tau_g$ be the toggle rate of cell $g$, which represents the average switching activity of the cell. In this work, we simplify the toggle rate factor by assuming that all gate cells have the same toggle rate, $\tau_g=\tau$, to reduce computational complexity and focus on the primary design trade-offs. Denote by $c_{s,g}(x,y,k,n)$ the count of cell $g\in\mathcal{G}$ in sub-block $s$ induced by the chosen format and by $n$ from Eq.(\ref{eq: aligner_width}). The MAC area and energy are
\begin{equation}
\text{Area}_{\text{MAC}} \;=\; \sum_{s\in\mathcal{S}}\sum_{g\in\mathcal{G}} c_{s,g}(x,y,k,n)\,A_g,\qquad
\text{Energy}_{\text{MAC}} \;=\; \sum_{s\in\mathcal{S}}\sum_{g\in\mathcal{G}} c_{s,g}(x,y,k,n)\,E_g \tau_g.
\end{equation}

\textbf{FP32 accumulator model.}
We model the FP32 accumulator by its combinational logic counts $c^{\text{ACC32}}_{g}$, yielding
\begin{equation}
\text{Area}_{\text{ACC32}} \;=\; \sum_{g\in\mathcal{G}} c^{\text{ACC32}}_{g}\,A_g,\qquad
\text{Energy}_{\text{ACC32}} \;=\; \sum_{g\in\mathcal{G}} c^{\text{ACC32}}_{g}\,E_g\tau_g.
\end{equation}

\textbf{Dequantizer model.}
We model the shared dequantizer based on the logic required for the specific format (e.g., fused integer additions or floating-point multiplications as described in \S\ref{sec:hardware_model_cost}). We aggregate its combinational logic counts $c^{\text{DEQ}}_{g}$, yielding
\begin{equation}
\text{Area}_{\text{DEQ}} \;=\; \sum_{g\in\mathcal{G}} c^{\text{DEQ}}_{g}\,A_g,\qquad
\text{Energy}_{\text{DEQ}} \;=\; \sum_{g\in\mathcal{G}} c^{\text{DEQ}}_{g}\,E_g\tau_g.
\end{equation}

\textbf{Total cost and per-lane reporting.}
The total MMU cost is
% 需要 \usepackage{amsmath}
\begin{equation}
\begin{split}
    \text{Area}_{\text{MMU}}   &= \text{Area}_{\text{MAC}} + \text{Area}_{\text{DEQ}} + \text{Area}_{\text{ACC32}}, \\
    \text{Energy}_{\text{MMU}} &= \text{Energy}_{\text{MAC}} + \text{Energy}_{\text{DEQ}} + \text{Energy}_{\text{ACC32}},
\end{split}
\end{equation}
and, when we report per-lane figures, we divide the cost of shared blocks (the dequantizer and the FP32 accumulator) by $k$.

\textbf{Summary.}
The hardware model includes the MAC unit, the dequantizer, and the FP32 accumulator; the quantizer is excluded from the overhead calculation. Given a low-precision format with exponent/mantissa widths $(x,y)$ (with $x{=}0$ for INT), a MAC array size $k$, an aligner cap $\texttt{psum\_bit\_width}$ (setting $n$ via Eq~(\ref{eq: aligner_width}), and technology cell factors $\{A_g,E_g\}_{g\in\mathcal{G}}$ (plus the dequantizer and FP32-accumulator gate counts), the model predicts the area and energy of the MAC and accumulation stages. It captures the relative cost trends across MX/NV-INT/FP formats at the same nominal bit width, the sensitivity to the aligner width $n$ (critical for FP addition), and the effect of sharing both the normalizer, the dequantizer, and the FP32 accumulator across $k$ lanes.

\section{More Details for Reproduction}~\label{sec:reproduction}

\subsection{Used Models}
\begin{table}[!h]
\centering
\caption{Huggingface IDs of evaluation models in direct-cast inference.}\label{tab:inference_model}
\begin{tabular}{cc}
\hline
Model Name      & Huggingface ID                    \\
\hline
Qwen3-0.6B      & Qwen/Qwen3-0.6B-Base              \\
Qwen3-1.7B      & Qwen/Qwen3-1.7B-Base              \\
Qwen3-4B        & Qwen/Qwen3-4B-Base                \\
Qwen3-8B        & Qwen/Qwen3-8B-Base                \\
Qwen3-14B       & Qwen/Qwen3-14B-Base               \\
Qwen3-32B       & Qwen/Qwen3-32B                    \\
Qwen3-30B-A3B   & Qwen/Qwen3-30B-A3B-Instruct-2507  \\
Qwen3-235B-A22B & Qwen/Qwen3-235B-22B-Instruct-2507 \\
Llama-3.2-1B    & meta-llama/Llama-3.2-1B           \\
Llama-3.2-3B    & meta-llama/Llama-3.2-3B           \\
Llama-3.1-8B    & meta-llama/Meta-Llama-3.1-8B      \\
Llama-3.1-70B   & meta-llama/Meta-Llama-3.1-70B    \\
\hline
\end{tabular}
\end{table}

\textbf{Models for inference evaluation.} We list the Huggingface IDs of evaluated open-sourced model for better reproduction in Tabel~\ref{tab:inference_model}. Note that we firstly choose the base model without supervise fine-tuning if it is open-sourced. For a model of a certain size, our selection principle is that if the base model is open source, we will first choose the base model; otherwise, we will select the model that has undergone SFT.

\begin{table}[!ht]
    \centering
    \caption{Llama-3 style Model architecture and training hyper-parameters.}
    \vspace{0.5em}
    \begin{tabular}{cccc}
    \toprule
    Model Size & 145M & 1B & 3B \\
    \midrule
    Layers & 12 & 16 & 28 \\
    Hidden Size & 1024 & 2048 & 3072 \\
    FFN Hidden Size & 3072 & 8192 & 8192\\
    Attention Heads & 16 & 32 & 24 \\
    KV Heads & 4 & 8 & 8\\
    \midrule
    Batch Size (\# Sequence) & 256 & 512 & 512 \\
    Max LR & 1.0e-3 & 6e-4 & 6e-4 \\
    Min LR & \multicolumn{3}{c}{0.1 $\times$ Max LR} \\
    Optimizer &  \multicolumn{3}{c}{AdamW ($\beta_1=0.9, \beta_2=0.95$)} \\ 
    Weight Decay &  \multicolumn{3}{c}{0.1} \\
    Clip Grad Norm & \multicolumn{3}{c}{1.0} \\
    LR Schedule & \multicolumn{3}{c}{Cosine} \\
    Warmup Steps & \multicolumn{3}{c}{500} \\
    Sequence Length & \multicolumn{3}{c}{2048} \\ 
    \bottomrule
    \end{tabular}
    \label{tab:model_settings}
\end{table}

\textbf{Models for training evaluation.} We select the Llama-3~\citep{llama3} style model for our experiments due to its wide adoption.  The Llama-3 style model employs Group Query Attention (GQA)\citep{gqa} for the self-attention module and SwiGLU\citep{swiglu} for the feed-forward module. Table~\ref{tab:model_settings} presents the detailed architectural settings and training hyper-parameters of the models used.

% \subsection{PROOF OF THEOREM}

\begin{table}[]
\centering
\caption{Ablation studies about the clipping range on INT8 quantization across quantization granularities, as well as  BFloat16 and UE8M0 scale factors. We report the 8-bit training loss (lower is better) on a 145M model with 20B training tokens. The baseline of BF16 training without quantization }\label{tab:int8_range_abl}
\begin{tabular}{ccc|cc}
\hline
            & \multicolumn{2}{c}{BF16 scale}             & \multicolumn{2}{c}{UE8M0 scale}             \\
\hline
            & {[}-128, 127{]} & \textbf{{[}-127, 127{]}} & {[}-128, 127{]} & \textbf{{[}-127, 127{]}} \\
\hline
per-channel & \textbf{3.2544} & 3.2560           & \textbf{3.3602} & \textbf{3.4307}          \\
256         & 3.1340  & \textbf{3.1307}          & 3.1628          & \textbf{3.1574}          \\
128         & 3.1309          & \textbf{3.1289}          & 3.1353          & \textbf{3.1326}          \\
64          & 3.1312 & \textbf{3.1269}          & 3.1312          & \textbf{3.1288}          \\
32          & 3.1354          & \textbf{3.1251}          & 3.1299          & \textbf{3.1269}    \\
\hline
\end{tabular}
\end{table}
\subsection{Necessity of Symmetric Integer Representation}\label{sec:appendix_sym_clip}
Table~\ref{tab:int8_range_abl} offer the ablation studies on representation range of INT8 quantization. We find that the bias in representation range would consistently degenerate INT8 training loss. For BFloat16 scale factor, we can find that asymmetric representation range even making block 32 quantization worse than block 256 quantization. This is because only the minimal values in each quantization block have possibility to be quantized into 128 in INT8 quantization, and smaller block size indicates more individual quantization blocks.  Additionally, asymmetric quantization also causes degeneration for UE8M0 scale factors, but the degeneration strength is slighter than BFloat16 scales. This is because UE8M0 scale factor consistently greater than or equal to Bfloat16 scale, leading less high-precision number to map to $Q_{min}$. These experiments demonstrate the necessity of symmetric representation space for integer quantization.

\begin{algorithm}[H]
    \caption{Analyzing Numerical Stability of Different Floating-Point Precisions}~\label{alg:precision}
    \label{alg:float_precision}
    \begin{algorithmic}[1]
        \State \textbf{Input:} Dimension $N=4096$, precision list $P = \{\text{bfloat16}, \text{float16}, \text{float32}\}$
        \State \textbf{Output:} Ratio of elements equal to 128 for each precision
        \For{each \textit{precision} in $P$}
            \State $D \gets \text{GenerateRandomMatrix}(N, N, \text{precision})$ \Comment{Generate $N \times N$ matrix from $\mathcal{N}(0, 1)$ on GPU}
            \State $S \gets D / 127$ \Comment{Calculate the scaler matrix}
            \State $D_{\text{norm}} \gets \text{Round}(D \oslash S)$ \Comment{$\oslash$ denotes element-wise division}
            \State $count \gets \text{CountElementsEqualTo}(D_{\text{norm}}, 128)$
            \State $total \gets N \times N$
            \State $ratio \gets count / total$
            \State \textbf{print} "Precision:", \textit{precision}, ", Ratio:", $ratio$
        \EndFor
    \end{algorithmic}
\end{algorithm}

\begin{table}[!h]
    \centering
    \caption{Results of Algorithm~\ref{alg:precision}.}
    \label{tab:precision_effect}
    \begin{tabular}{ccc}
    \hline
     BFloat16 & Float16 & Float32  \\
    \hline
    16.82\% & 0.02\% & 0 \\
    \hline
    \end{tabular}
\end{table}
\textbf{Numerical stability analysis.}
We also analyze the numerical stability of different float-point for quantization mapping through Algorithm~\ref{alg:precision}. Table~\ref{tab:precision_effect} shows the results of  Algorithm~\ref{alg:precision}, demonstrating that in BFloat16 precision,  a significant portion of values (16.82\%) are mapped to -128. This phenomenon occurs even though the scaling factor s is theoretically designed to map the value to 127. In conclusion, this analysis highlights a critical pitfall of using low-precision floating-point formats for quantization calculations. The inherent lack of precision in bfloat16 and, to a lesser extent, float16 can lead to overflow during the scaling step, incorrectly mapping values to outside the intended integer range. This powerfully demonstrates that a forced symmetric clipping step is essential for guaranteeing the correctness and stability of quantization, particularly when the computation is performed using low-precision data types.

\subsection{Detailed Results}\label{sec:detailed_results}
This section offer detailed numbers of experiments, as follows:
\begin{itemize}
    \item Table~\ref{tab:qwen_kl} and Table~\ref{tab:llama_kl} present the KL divergence results, corresponding to Table~\ref{tab:inference_summary}. 
    \item Table~\ref{tab:qwen_ppl} and Table~\ref{tab:llama_ppl} present the perplexity results, for better understanding the relationship between KL divergence and perplexity. They are consistent in most case.
\end{itemize}

\begin{table}[!ht]
\centering
% \caption{Qwen3 models KL (lower is better) results across different low-bit formats in direct-cast inference.}
\caption{Qwen3 models KL divergence (lower is better) results across different low-bit formats in direct-cast inference. All reported KL metrics are the average over all tokens, multiplied by $10^6$.}
\label{tab:qwen_kl}
\begin{tabular}{ccccccccc}
\hline
       & \multicolumn{8}{c}{Qwen-3}                                                                                                          \\
\hline
Format & 0.6B           & 1.7B           & 4B             & 8B             & 14B            & 32B            & 30B-A3B       & 235B-A22B     \\
\hline
MXINT8 & \textbf{191}   & \textbf{209}   & \textbf{112}  & \textbf{168}  & \textbf{96}   & \textbf{118}   & \textbf{160}  & \textbf{276}   \\
MXFP8  & 579            & 406            & 346           & 362           & 300           & 457            & 380           & 483            \\
\cdashline{1-9}
MXINT6 & 1944           & 2464           & 928           & 1104          & 804           & 1012           & 768           & 1333           \\
MXFP6  & \textbf{1030}  & \textbf{874}   & \textbf{539}  & \textbf{592}  & \textbf{467}  & \textbf{627}   & \textbf{606}  & \textbf{1099}  \\
\cdashline{1-9}
MXINT4 & 39936          & 30208          & 17408         & 15552         & 34304         & 27392          & 13248         & 16331          \\
MXFP4  & \textbf{17602} & \textbf{14614} & \textbf{8568} & \textbf{8228} & \textbf{8119} & \textbf{10302} & \textbf{6194} & \textbf{16238} \\
\cdashline{1-9}
NVINT4 & 10560          & 8320           & 4864          & 5120          & 5568          & 7968           & 3120          & 9702           \\
NVFP4  & \textbf{8104}  & \textbf{4995}  & \textbf{3844} & \textbf{3430} & \textbf{2835} & \textbf{3778}  & \textbf{2443} & \textbf{9238} \\
\cdashline{1-9}
\hline
       & \multicolumn{8}{c}{Qwen-3 \textbf{(w/ random Hadamard rotation)}}                                         \\
\hline
Format & 0.6B           & 1.7B           & 4B             & 8B             & 14B           & 32B           & 30B-A3B       & 235B-A22B     \\
\hline
MXINT8 & \textbf{137}   & \textbf{150}   & \textbf{80}   & \textbf{130}  & \textbf{70}   & \textbf{88}   & \textbf{135}  & \textbf{229}  \\
MXFP8  & 921            & 1321           & 468           & 577           & 393           & 497           & 391           & 707           \\
\cdashline{1-9}
MXINT6 & 1137           & \textbf{1274}  & 547           & 690           & 481           & 615           & 444           & 809           \\
MXFP6  & \textbf{1007}  & 1446  & \textbf{497}  & \textbf{618}  & \textbf{454}  & \textbf{558}  & \textbf{422}  & \textbf{740}  \\
\cdashline{1-9}
MXINT4 & 26488          & 26578          & 10498         & 12241         & 8459          & 9510          & 6080          & 9660          \\
MXFP4  & \textbf{17995} & \textbf{20443} & \textbf{7260} & \textbf{8562} & \textbf{6410} & \textbf{6536} & \textbf{5087} & \textbf{7058} \\
\cdashline{1-9}
NVINT4 & \textbf{7771}  & \textbf{7236}  & \textbf{3431} & \textbf{4026} & \textbf{3070} & \textbf{3647} & \textbf{2222} & \textbf{3931} \\
NVFP4  & 12031          & 10582          & 5065          & 5912          & 4214          & 4662          & 3200          &   5786   \\   
\hline
\end{tabular}
\end{table}
\begin{table}[!h]
\centering
% \caption{Llama-3 models KL (lower is better) results across different low-bit formats in direct-cast inference.}
\caption{Llama-3 models KL divergence (lower is better) results across different low-bit formats in direct-cast inference. All reported KL metrics are the average over all tokens, multiplied by $10^6$.}\label{tab:llama_kl}
\begin{tabular}{ccccc}
\hline
                   & \multicolumn{4}{c}{Llama}                                         \\
                   \hline
Format             & 3.2-1B         & 3.2-3B         & 3.1-8b         & 3.1-70B        \\
\hline
MXINT8 & \textbf{111}    & \textbf{77}     & \textbf{82}     & \textbf{191}     \\
MXFP8  & 464             & 325             & 359             & 514              \\
\cdashline{1-5}
MXINT6 & 1133            & 743             & 776             & 1744             \\
MXFP6  & \textbf{651}    & \textbf{457}    & \textbf{491}    & \textbf{1436}    \\
\cdashline{1-5}
MXINT4 & 26153           & 14089           & 12380           & 22538            \\
MXFP4  & \textbf{14446}  & \textbf{8251}   & \textbf{7586}   & \textbf{21372}   \\
\cdashline{1-5}
NVINT4 & 7508            & 4312            & 4224            & 10970            \\
NVFP4  & \textbf{5691}   & \textbf{3684}   & \textbf{3718}   & \textbf{10544}  \\
\hline
       & \multicolumn{4}{c}{Llama\textbf{(w/ random Hadamard rotation)}}                                           \\
       \hline
Format & 3.2-1B         & 3.2-3B         & 3.1-8b         & 3.1-70B         \\
\hline
MXINT8 & \textbf{89}    & \textbf{63}   & \textbf{65}   & \textbf{145}    \\
MXFP8  & 573            & 388           & 409           & 1393            \\
\cdashline{1-5}
MXINT6 & 773            & 531           & 558           & 1518            \\
MXFP6  & \textbf{643}   & \textbf{447}  & \textbf{457}  & \textbf{1476}   \\
\cdashline{1-5}
MXINT4 & 20126          & 11116         & 10272         & 137612          \\
MXFP4  & \textbf{11967} & \textbf{8269} & \textbf{7189} & \textbf{129471} \\
\cdashline{1-5}
NVINT4 & \textbf{5854}  & \textbf{3912} & \textbf{3609} & \textbf{19975}  \\
NVFP4  & 8129           & 5240          & 4752          & 77363    \\
\hline
\end{tabular}
\end{table}

\begin{table}[!h]
\centering
 \setlength{\tabcolsep}{4pt}
\caption{Qwen3 models perplexity (lower is better) results  of WikiText2 across different low-bit formats in direct-cast inference.}\label{tab:qwen_ppl}
\begin{tabular}{ccccccccc}
\hline
       & \multicolumn{8}{c}{Qwen-3}                                                                                                                                 \\
\hline
Format & 0.6B             & 1.7B                       & 4B               & 8B              & 14B             & 32B             & 30B-A3B         & 235B-A22B       \\
\hline
BF16   & 11.5868          & \multicolumn{1}{l}{8.7084} & 7.3368           & 6.5135          & 5.9498          & 7.0168          & 6.8178          & 4.0929      \\
\hline
MXINT8 & \textbf{11.6377} & \textbf{8.7424}            & \textbf{7.3511} & \textbf{6.5174} & \textbf{5.955}  & \textbf{7.0185} & \textbf{6.8167} & \textbf{4.0959} \\
MXFP8  & 11.7494          & 8.7822                     & 7.3813          & 6.5444          & 5.9711          & 7.0357          & 6.8335          & 4.1101          \\
\cdashline{1-9}
MXINT6 & 12.2297          & 9.2622                     & 7.496           & 6.6499          & 6.0483          & 7.05            & 6.8745          & 4.1743          \\
MXFP6  & \textbf{11.9108} & \textbf{8.8961}            & \textbf{7.4135} & \textbf{6.5825} & \textbf{5.9953} & \textbf{7.0285} & \textbf{6.8467} & \textbf{4.1662} \\
\cdashline{1-9}
MXINT4 & 48.6713          & 21.8749                    & 11.9487         & 10.0423         & 16.7227         & 15.1619         & 9.3837          & 5.918           \\
MXFP4  & \textbf{20.4522} & \textbf{24.0766}           & \textbf{9.1553} & \textbf{8.0135} & \textbf{7.2471} & \textbf{8.2047} & \textbf{7.8203} & \textbf{5.9007} \\
\cdashline{1-9}
NVINT4 & 15.9729          & 10.9128                    & 8.3304          & 7.415           & 6.81            & 8.0161          & 7.2024          & 4.8916          \\
NVFP4  & \textbf{14.6818} & \textbf{9.9966}            & \textbf{8.0144} & \textbf{7.0285} & \textbf{6.3129} & \textbf{7.3604} & \textbf{7.1874} & \textbf{4.8309} \\
\hline
       & \multicolumn{8}{c}{Qwen-3\textbf{(w/ random Hadamard rotation)}}                                                                                                                      \\
       \hline
Format & 0.6B             & 1.7B             & 4B              & 8B              & 14B             & 32B             & 30B-A3B         & 235B-A22B       \\
\hline
MXINT8               & \textbf{11.6179}     & \textbf{8.7240}      & \textbf{7.3407}      & \textbf{6.5170}      & \textbf{5.9521}      & \textbf{7.0187}      & \textbf{6.8231}      & \textbf{4.0973}      \\
MXFP8                & 11.8629              & 8.9972               & 7.4068               & 6.5898               & 5.9839               & 7.0448               & 6.8918               & 4.1287               \\
\cdashline{1-9}
MXINT6               & \textbf{11.9422}     & \textbf{9.0122}      & \textbf{7.4071}      & 6.6119               & 5.9905               & \textbf{7.0627}      & \textbf{6.8666}      & \textbf{4.1263}      \\
MXFP6                & 11.9096              & 9.0089               & 7.4108               & \textbf{6.5911}      & \textbf{5.9981}      & 7.0787               & 6.8711               & 4.1252               \\
\cdashline{1-9}
MXINT4               & \textbf{28.6510}     & \textbf{21.3032}     & \textbf{9.8238}      & \textbf{9.2029}      & \textbf{7.3564}      & \textbf{8.2083}      & \textbf{7.8292}      & \textbf{4.9891}      \\
MXFP4                & 20.3684              & 15.9527              & 8.8148               & 8.1113               & 6.9521               & 7.7401               & 7.9673               & 4.7035               \\
\cdashline{1-9}
NVINT4               & \textbf{14.6052}     & \textbf{10.7822}     & \textbf{7.9824}      & \textbf{7.1705}      & \textbf{6.3702}      & \textbf{7.3625}      & \textbf{7.1557}      & \textbf{4.3913}      \\
NVFP4  & 16.5762          & 11.7541          & 8.2716          & 7.5084          & 6.5427          & 7.4522          & 7.3214          & 4.5918 \\
\hline
\end{tabular}
\end{table}
\begin{table}[!h]
\centering
\caption{Llama-3 models perplexity (lower is better) results of WikiText2 across different low-bit formats in direct-cast inference.}\label{tab:llama_ppl}
\begin{tabular}{ccccc}
\hline
       & \multicolumn{4}{c}{Llama}                                               \\
\hline
Format & 3.2-1B           & 3.2-3B           & 3.1-8b          & 3.1-70B         \\
\hline
BF16   & \textbf{9.0625}  & \textbf{7.2857}  & \textbf{5.8402} & \textbf{2.637}  \\
\cdashline{1-5}
MXINT8 & \textbf{9.0815}  & \textbf{7.2944} & \textbf{5.8487} & \textbf{2.6674} \\
MXFP8  & 9.1695           & 7.3381          & 5.895           & 2.6674          \\
\cdashline{1-5}
MXINT6 & 9.3557           & 7.4184          & 5.9643          & 2.7298          \\
MXFP6  & \textbf{9.2209}  & \textbf{7.3605} & \textbf{5.916}  & \textbf{2.7298} \\
\cdashline{1-5}
MXINT4 & 21.9893          & 11.2715         & 8.7408          & 5.1894          \\
MXFP4  & \textbf{14.0516} & \textbf{9.2355} & \textbf{6.4845} & \textbf{4.9492} \\
\cdashline{1-5}
NVINT4 & 11.3987          & 8.225           & 6.5957          & 3.5502          \\
NVFP4  & \textbf{10.7473} & \textbf{8.0343} & \textbf{6.4917} & \textbf{3.492}  \\
\hline
       & \multicolumn{4}{c}{Llama\textbf{(w/ random Hadamard rotation)}}                                                  \\
        \hline
Format & 3.2-1B           & 3.2-3B           & 3.1-8b          & 3.1-70B            \\
\hline
MXINT8 & \textbf{9.0715}  & \textbf{7.2912} & \textbf{5.845}  & \textbf{2.6428}    \\
MXFP8  & 9.1932           & 7.3465          & 5.9001          & 2.7232             \\
\cdashline{1-5}
MXINT6 & 9.2622           & 7.3828          & 5.9276          & 2.7333             \\
MXFP6  & \textbf{9.2204}  & \textbf{7.3703} & \textbf{5.9075} & \textbf{2.735}     \\
\cdashline{1-5}
MXINT4 & 17.9797          & 10.3057         & 8.0745          & 1146.7256          \\
MXFP4  & \textbf{13.3987} & \textbf{9.262}  & \textbf{7.2318} & \textbf{1118.4431} \\
\cdashline{1-5}
NVINT4 & \textbf{10.8399} & \textbf{8.1119} & \textbf{6.4701} & \textbf{4.9786}    \\
NVFP4  & 11.7635          & 8.4693          & 6.7028          & 79.7586           \\
\hline
\end{tabular}
\end{table}

\end{document}